\documentclass[dvipsnames,format=sigconf]{acmart}

\AtBeginDocument{%
\providecommand\BibTeX{{%
\normalfont B\kern-0.5em{\scshape i\kern-0.25em b}\kern-0.8em\TeX}}}
    
\usepackage{xcolor}
\usepackage{colortbl}
\usepackage{booktabs}
\usepackage{amsmath}
\usepackage{algorithm}
\usepackage{algpseudocode}
\usepackage{multirow}
\usepackage{subcaption}
\usepackage{caption}
\usepackage{algorithm}
\usepackage{algpseudocode}

\usepackage{placeins}

\algtext*{EndFor}
\algtext*{EndIf}
\algtext*{EndWhile}
\algtext*{EndFunction}
\algtext*{EndProcedure}

\usepackage{booktabs}
\acmDOI{10.1145/nnnnnnn.nnnnnnn} % To be updated after completing copyright process
\acmISBN{978-x-xxxx-xxxx-x/YY/MM} % To be updated after completing copyright process
\acmConference[GECCO '26]{The Genetic and Evolutionary Computation Conference 2026}{July 13--17, 2026}{San José, Costa Rica}
\acmYear{2026}
\copyrightyear{2026}

\begin{document}

\title{Cooperative Coevolution versus Monolithic Evolutionary Search for Semi-Supervised Tabular Classification}

\author{Jamal Toutouh}
\affiliation{
\institution{ITIS, Universidad de Málaga}
\country{Spain}
}
\email{jamal@uma.es}

\begin{abstract}
%Semi-supervised learning (SSL) for tabular classification remains practically important but methodologically brittle: classical self-labeling pipelines can collapse under threshold miscalibration, class imbalance, and confirmation bias, while modern deep tabular SSL often increases implementation and computational complexity. We propose \emph{CC-SSL}, a cooperative coevolutionary framework that jointly adapts (i) a \emph{view builder} that induces complementary feature views from tabular predictors and (ii) a \emph{pseudo-labeling policy} that governs which unlabeled samples are accepted, when, and under what balancing constraints. CC-SSL is designed to be lightweight: it uses classical learners (e.g., logistic regression, linear SVM, random forests) and a robust, distribution-aware fitness that aggregates performance over multiple resampled labeled splits while penalizing instability and overly aggressive pseudo-label additions. We detail the encoding, variation operators, and the coevolutionary evaluation protocol, and we outline an OpenML-based experimental plan with rigorous statistical testing (Wilcoxon with Holm--Bonferroni correction and effect sizes). 

This paper studies semi-supervised tabular classification in the extreme low-label regime using lightweight base learners. The paper proposes a cooperative coevolutionary method (CC-SSL) that evolves (i) two feature-subset views and (ii) a pseudo-labeling policy, and compares it to a matched monolithic evolutionary baseline (EA-SSL) and three lightweight SSL baselines. Experiments on 25 OpenML datasets with labeled fractions $\{1\%,5\%,10\%\}$ evaluate test MacroF1 and accuracy, together with evolutionary and pseudo-label diagnostics. CC-SSL and EA-SSL achieve higher median test MacroF1 than the lightweight baselines, with the largest separations at $1\%$ labeled data. Most CC-SSL vs.\ EA-SSL comparisons are statistical draws on final test performance. EA-SSL shows higher best-so-far fitness and higher diversity during search, while time-to-target is comparable and generations-to-target favors EA-SSL in several multiclass settings. Pseudo-label volume, ProbeDrop, and validation optimism show no significant differences between CC-SSL and EA-SSL under the shared protocol.
\end{abstract}




\maketitle

\setlength{\textfloatsep}{8pt plus 2pt minus 2pt}   % between float and text

\section{Introduction}
\label{sec:intro}

Tabular classification tasks often operate in a low-label regime, where feature vectors are available but labels are scarce. Semi-supervised learning (SSL) targets this setting by combining labeled and unlabeled data, but SSL can degrade performance when method assumptions are violated or when early pseudo-label errors dominate training \cite{chapelle2006ssl,vanengelenhoos2020survey,singh2008unlabeled}.

Lightweight SSL for tabular data typically wraps supervised learners with pseudo-labeling, multi-view exchange, or graph-based propagation. These approaches are sensitive to view construction, confidence calibration, and graph design choices \cite{triguero2015selflabeled,blum1998cotraining,zhu2002labelprop}. Neural tabular SSL methods can be effective but increase architecture and tuning complexity, whereas large benchmarks report that tree-based and other classical models remain strong competitors on tabular problems \cite{grinsztajn2022trees}.

This paper uses evolutionary search to optimize coupled SSL design decisions that are difficult to tune manually in wrapper-style pipelines. The search targets two interacting components: (i) a view builder that selects complementary feature subspaces and (ii) a pseudo-labeling policy that controls pseudo-label selection and stopping behavior. The paper evaluates whether factorizing this joint search with cooperative coevolution (CC) improves outcomes relative to a monolithic evolutionary algorithm (EA) that encodes both components in a single genotype \cite{eiben2015ec,potter1994ccga}.

The paper compares two matched evolutionary SSL methods for tabular classification: \textbf{CC-SSL}, which evolves the view builder and pseudo-labeling policy in cooperating populations, and \textbf{EA-SSL}, which evolves the same components jointly in a single population. Experiments on OpenML datasets with labeled fractions $\{1\%,5\%,10\%\}$ compare both methods against lightweight baselines (Self-Training, Heuristic Co-Training, and Label Spreading) using MacroF1 score, accuracy, and behavioral and generalization metrics.

The empirical analysis addresses the following research questions: 
\textbf{[RQ1]:} Do evolutionary SSL methods outperform lightweight SSL baselines in the extreme low-label regime?,
\textbf{[RQ2]:} When does CC improve SSL relative to monolithic evolutionary search under matched budgets?
\textbf{[RQ3]:} When CC-SSL and EA-SSL differ in performance, do these differences co-occur with changes in pseudo-label dynamics, stability proxies, or validation--test optimism?, 
\textbf{[RQ4]:} Is either approach preferable under a compute--performance trade-off measured with a runtime proxy?.

The paper makes the following contributions:
(i) a joint optimization formulation for lightweight tabular SSL that couples two-view feature-subset construction with pseudo-labeling policy design;
(ii) a controlled CC vs.\ EA comparison using shared building blocks;
(iii) a staged tuning protocol that restricts operator and fitness-weight selection to a small development set; and
(iv) a benchmark evaluation against lightweight SSL baselines using multi-metric reporting and reproducible artifacts.

The paper is organized as follows:
Section~\ref{sec:related-work} reviews SSL for tabular classification; Section~\ref{sec:method} describes CC-SSL and EA-SSL; Section~\ref{sec:exp_settings} describes the experimental settings; Section~\ref{sec:experimental_results} presents results and answers the research questions; and Section~\ref{sec:conclusions} draws the main conclusions and future work.

\section{Related Work}
\label{sec:related-work}

\sloppy 
Let $\mathcal{D}=\mathcal{D}_{\ell}\cup\mathcal{D}_{u}$ denote a dataset with $n_{\ell}$ labeled samples \hbox{$\mathcal{D}_{\ell}=\{(x_i,y_i)\}_{i=1}^{n_{\ell}}$} and $n_u$ unlabeled samples $\mathcal{D}_{u}=\{x_j\}_{j=1}^{n_u}$, where $x\in\mathbb{R}^{d}$ after preprocessing and $y\in\{1,\dots,C\}$.
SSL learns a classifier $f_{\theta}$ by exploiting $\mathcal{D}_{\ell}$ and $\mathcal{D}_{u}$ to reduce expected risk on unseen samples~\cite{vanengelenhoos2020survey}.
Unlabeled data is not guaranteed to help, and performance can degrade when method assumptions are violated or when incorrect pseudo-labels dominate training~\cite{singh2008unlabeled}.
We review lightweight SSL for tabular classification and summarizes CC mechanisms that motivate factorized vs. monolithic evolutionary search.

\sloppy
%\paragraph{Wrapper-style SSL (self-training and multi-view variants).}Self-training fits a supervised model on $\mathcal{D}_{\ell}$, predicts labels for $\mathcal{D}_{u}$, and adds a subset of high-confidence pseudo-labeled samples back into the labeled pool.Large empirical studies report sensitivity to thresholding, stopping criteria, calibration, and class imbalance~\cite{triguero2015selflabeled,vanengelenhoos2020survey}. Co-training exchanges pseudo-labels across classifiers trained on different views~\cite{blum1998cotraining}, while tri-training uses three classifiers and agreement-based selection to reduce dependence on strict view assumptions~\cite{zhou2005tritraining}. Tabular data rarely provides natural multi-view structure, which makes view construction (feature splits, projections, masks) a central design choice that interacts with pseudo-label policies.

\sloppy
\paragraph{Wrapper-style SSL (self-training and multi-view variants)}
Self-training fits a supervised model on $\mathcal{D}_{\ell}$, predicts labels for $\mathcal{D}_{u}$, and adds high-confidence pseudo-labeled samples to the labeled pool.
Empirical studies report sensitivity to thresholding, stopping criteria, calibration, and class imbalance~\cite{triguero2015selflabeled,vanengelenhoos2020survey}.
Co-training exchanges pseudo-labels across classifiers trained on different views~\cite{blum1998cotraining}, while tri-training uses agreement-based selection to reduce dependence on view assumptions~\cite{zhou2005tritraining}.
Tabular data rarely provides natural multi-view structure, which makes view construction a central design choice that interacts with pseudo-label policies.

\paragraph{Graph-based SSL}
Graph-based methods build a similarity graph and propagate labels along edges, exploiting cluster or manifold assumptions.
Label propagation is a canonical approach~\cite{zhu2002labelprop}, and local-and-global consistency provides a regularized variant~\cite{zhou2003llgc}.
In heterogeneous tabular data, performance depends on preprocessing and graph construction, and computational cost grows with $n_u$.

%\paragraph{Graph-based SSL.}
%Graph-based methods build a similarity graph over samples and propagate labels along edges, exploiting cluster or manifold assumptions.
%Label propagation is a canonical approach~\cite{zhu2002labelprop}, and the local-and-global consistency formulation provides a widely used regularized variant~\cite{zhou2003llgc}.
%In heterogeneous tabular spaces, performance depends on preprocessing and graph construction (feature scaling, distance metric, neighborhood size), and computational cost grows quickly with $n_u$.

\paragraph{Modern tabular SSL and representation learning}
Recent methods adapt self-supervision and contrastive learning to tabular inputs.
VIME combines feature corruption with mask estimation and semi-supervised fine-tuning~\cite{yoon2020vime}. SCARF constructs contrastive views via random feature corruption~\cite{bahri2021scarf}.
Transformer-based approaches use attention and contrastive objectives for low-label regimes~\cite{huang2020tabtransformer,somepalli2021saint}.
These methods can improve performance but typically require neural architectures and additional tuning, while strong non-neural baselines remain competitive across diverse tabular datasets~\cite{grinsztajn2022trees}.

%\paragraph{Modern tabular SSL and representation learning.}Recent methods adapt self-supervised and contrastive learning to tabular inputs.VIME combines feature corruption with a mask-estimation objective and semi-supervised fine-tuning~\cite{yoon2020vime}. SCARF creates contrastive views via random feature corruption~\cite{bahri2021scarf}. Transformer-based approaches incorporate contextual embeddings, attention, and pretraining or contrastive objectives for label-scarce regimes~\cite{huang2020tabtransformer,somepalli2021saint}. These methods can improve performance but typically require neural architectures and additional tuning. Benchmarking results indicate that strong non-neural baselines remain competitive across diverse tabular datasets~\cite{grinsztajn2022trees}, which motivates lightweight approaches when engineering and tuning budgets are limited.

\paragraph{Cooperative coevolution and collaboration strategies}
CC decomposes a solution into interacting components that evolve in separate populations and are evaluated jointly~\cite{potter1994ccga}.
Collaboration choices (e.g., partner selection, archives) can affect stability and search dynamics in cooperative settings~\cite{panait2005comal}.
These results motivate treating factorization vs. monolithic encoding as a primary design axis when comparing CC against standard evolutionary search.

\paragraph{Gap and connection to this paper}
Prior lightweight tabular SSL methods require coupled decisions about view construction and pseudo-label acceptance, and performance can be unstable when $n_{\ell}$ is very small~\cite{triguero2015selflabeled,vanengelenhoos2020survey}.
Prior cooperative coevolution work establishes general principles for decomposition and collaboration~\cite{potter1994ccga,panait2005comal}, but it does not provide a controlled comparison against a matched monolithic EA in evolutionary tabular SSL with shared building blocks.
This paper treats view construction and pseudo-label policy as two co-adapted components and compares CC-SSL against EA-SSL under a matched representation budget.

\section{Evolutionary Pseudo-Labeling SSL for Tabular Classification}
\label{sec:method}

The proposed CC-SSL and EA-SSL methods are based on pseudo-labeling (self-training) SSL. 
Both methods share the same building blocks, evaluation protocol, and fitness definition; the methods differ only in how the search space is factorized.

\subsection{Pseudo-labeling SSL}
\label{sec:ssl_inner}

Pseudo-labeling is a wrapper-style SSL procedure that augments labeled data with confident predictions on unlabeled samples \cite{triguero2015selflabeled,vanengelenhoos2020survey}.
Let $L_0=\mathcal{D}_{\ell}$ and $U_0=\mathcal{D}_{u}$ denote the initial labeled and unlabeled sets.
At iteration $t\in\{0,\dots,T-1\}$, the procedure fits a supervised model on $L_t$, predicts class posteriors on $U_t$, accepts a subset of pseudo-labels that satisfy a policy, and updates $(L_{t+1},U_{t+1})$.
The procedure stops after $T$ iterations or when no samples are accepted.

The present work uses two views of the same tabular input to enable disagreement-based controls.
A view builder $a\in\mathcal{A}$ defines two mappings $\phi^{(1)}_a(\cdot)$ and $\phi^{(2)}_a(\cdot)$.
A policy $b\in\mathcal{B}$ specifies a base learner family $\mathcal{H}_b$ and acceptance rules.
At each iteration, two models $h^{(1)}_t,h^{(2)}_t\in\mathcal{H}_b$ are trained on the two views of $L_t$ and produce posterior vectors $p^{(1)}_t(\cdot\mid x)$ and $p^{(2)}_t(\cdot\mid x)$ for each $x\in U_t$.

%\paragraph{Acceptance rules.}
For $x\in U_t$, define $\hat{y}^{(v)}_t(x)=\arg\max_y p^{(v)}_t(y\mid x)$ and $c^{(v)}_t(x)=\max_y p^{(v)}_t(y\mid x)$.
The policy $b$ defines: (i) a confidence threshold schedule $\tau_t$; (ii) an optional margin constraint $\gamma$ based on the gap between the largest and second-largest posterior; (iii) an optional disagreement veto requiring $\hat{y}^{(1)}_t(x)=\hat{y}^{(2)}_t(x)$; (iv) per-class caps $q_t$ on accepted pseudo-labels per iteration; and (v) stopping criteria.
These controls target documented sensitivities of self-labeled SSL to thresholding, calibration, imbalance, and error amplification~\cite{triguero2015selflabeled}.

Algorithm~\ref{alg:pseudolabel} summarizes the SSL loop. The explicit procedure is reported because the interaction between schedules, veto rules, caps, and stopping criteria materially affects performance in low-label settings \cite{triguero2015selflabeled}.

\begin{algorithm}[h]
\caption{Pseudo-labeling SSL with two views (view builder $a$) and policy $b$}
\small
\label{alg:pseudolabel}
\begin{algorithmic}[1]
\Require $L_0, U_0$, view builder $a$, policy $b$, max iterations $T$
\For{$t=0,\dots,T-1$}
  \State Train $h^{(1)}_t\in\mathcal{H}_b$ on $\{(\phi^{(1)}_a(x),y):(x,y)\in L_t\}$
  \State Train $h^{(2)}_t\in\mathcal{H}_b$ on $\{(\phi^{(2)}_a(x),y):(x,y)\in L_t\}$
  \State Compute $p^{(v)}_t(\cdot\mid x)$ for all $x\in U_t$ and $v\in\{1,2\}$

  \State $C_t \gets U_t$
  \State Filter by confidence: $C_t \gets \{x\in C_t : \max_{v} c^{(v)}_t(x) \ge \tau_t\}$
  \If{margin is enabled}
    \State Filter by margin: $C_t \gets \{x\in C_t : \max_{v}\Delta^{(v)}_t(x) \ge \gamma\}$
  \EndIf
  \If{veto is enabled}
    \State Filter by agreement: $C_t \gets \{x\in C_t : \hat{y}^{(1)}_t(x)=\hat{y}^{(2)}_t(x)\}$
  \EndIf

  \State Select up to $q_{t,c}$ candidates per class from $C_t$ %(highest confidence first)
  \State Add selected pseudo-labeled samples to $L_{t+1}$ and remove them from $U_{t+1}$
  \If{no samples were added}
    \State \textbf{break}
  \EndIf
\EndFor
\State \Return $L_{\text{final}}$ and final models $\{h^{(v)}\}_{v=1}^{2}$
\end{algorithmic}
\end{algorithm}

\subsection{Search space and genotype representation}
\label{sec:search_space}

Both CC-SSL and EA-SSL optimize the same two component spaces.
A view builder $a\in\mathcal{A}$ defines two tabular views used by the inner SSL loop (Algorithm~\ref{alg:pseudolabel}).
A policy $b\in\mathcal{B}$ defines the base learner configuration and the pseudo-label acceptance rules that govern the evolution of $(L_t,U_t)$.
This subsection specifies the genotype of each population and its mapping to phenotypes used during evaluation.

Population $\mathcal{A}$ individuals (view builders) are encoded as \hbox{$a=\big(m^{(1)},m^{(2)},p^{(1)},p^{(2)},k^{(1)},k^{(2)},B^{(1)},B^{(2)}\big)$}, 
where each view $v\in\{1,2\}$ is defined by:
(i) a binary feature mask $m^{(v)}\in\{0,1\}^{d}$ with a minimum
number of selected features,
(ii) a projection flag $p^{(v)}\in\{0,1\}$,
(iii) a projection dimension $k^{(v)}\in\{2,\dots,k_{\max}\}$, and
(iv) a discretization bin parameter $B^{(v)}\in\{0,\dots,B_{\max}\}$. 
The representation assigns independent genes to the two views to allow overlap but does not enforce disjointness.

Population $\mathcal{B}$ individuals (pseudo-labeling policies)
are encoded as \hbox{$b=\big(\theta_{\mathrm{clf}},\tau_0,\Delta\tau,\tau_{\min},q,\gamma,\nu,T\big)$},
where $\theta_{\mathrm{clf}}$ denotes base learner hyperparameters (a regularized linear classifier),
$\tau_0\in(0,1)$ is the initial confidence threshold,
$\Delta\tau\ge 0$ is the per-iteration threshold decay,
$\tau_{\min}\in(0,1)$ is the lower bound on the threshold,
$q\in\mathbb{N}$ is the per-class cap on accepted pseudo-labels per iteration,
$\gamma\in[0,1]$ is the minimum probability margin,
$\nu\in\{0,1\}$ is the disagreement veto flag,
and $T\in\mathbb{N}$ is the maximum number of pseudo-labeling iterations.
The cap $q$ limits pseudo-label imbalance and pseudo-label volume, which affects error amplification in self-labeled SSL~\cite{triguero2015selflabeled}.

%CC-SSL evolves $a$ and $b$ in separate populations, with each component using the same encoding as its corresponding segment in EA-SSL. EA-SSL concatenates the two encodings into a single genotype representing a complete pair $(a,b)$. This construction matches the search degrees of freedom and isolates the effect of factorization and collaboration. Table~\ref{tab:genotype_spec} in the Appendix reports genotype domains and constraints.

CC-SSL evolves $a$ and $b$ in separate populations, with each component using the same encoding as its segment in EA-SSL. EA-SSL concatenates the two encodings into a genotype representing a pair $(a,b)$. This construction matches the degrees of freedom and isolates the effect of factorization and collaboration. Table~\ref{tab:genotype_spec} in the Appendix reports genotype domains and constraints.

\subsection{CC-SSL and EA-SSL methods}
\label{sec:ccssl}
%CC-SSL uses cooperative coevolution with two interacting populations, Pop-$\mathcal{A}=\{a_i\}_{i=1}^{N_A}$ and Pop-$\mathcal{B}=\{b_j\}_{j=1}^{N_B}$, and assigns fitness through collaboration using elitist and randomly sampled partners~\cite{potter1994ccga,Potter2000CoadaptedSubcomponents}. This subsection specifies the collaboration rule, the joint fitness evaluation, and the population update used in the experiments. Algorithm~\ref{alg:ccssl_compact} summarizes the procedure. Each population evolves independently via tournament selection and elitist generational replacement~\cite{eiben2015ec}, and variation operators (crossover, mutation, feasibility repair) are applied within each population.

CC-SSL uses cooperative coevolution with two interacting populations, Pop-$\mathcal{A}={a_i}{i=1}^{N_A}$ and Pop-$\mathcal{B}={b_j}{j=1}^{N_B}$, and assigns fitness through collaboration using elitist and randomly sampled partners~\cite{potter1994ccga,Potter2000CoadaptedSubcomponents}. This subsection specifies the collaboration rule, joint fitness evaluation, and population update. Algorithm~\ref{alg:ccssl_compact} summarizes the procedure. Each population evolves independently via tournament selection and elitist generational replacement~\cite{eiben2015ec}, and variation operators are applied within each population.

\noindent\textbf{Collaboration and credit assignment.}
Partner choice affects stability and credit assignment in cooperative settings~\cite{PanaitLuke2005}. CC-SSL assigns fitness using two collaborator types from the opposite population: the current best individual (\textit{elitist partner}) and $R$ uniformly sampled partners (\textit{random partners}). 

For $a_i\in$ Pop-$\mathcal{A}$,
The collaborators are defined as \hbox{$\mathcal{P}_B(a_i)=\{b^{\mathrm{best}}\}\cup\{b^{(r)}\}_{r=1}^{R}$} and assign a fitness value to $a_i$ as $\mathrm{fit}(a_i)=\max_{b\in \mathcal{P}_B(a_i)} F(a_i,b)$.
The symmetric rule assigns $\mathrm{fit}(b_j)=\max_{a\in \mathcal{P}_A(b_j)} F(a,b_j)$ for $b_j\in$ Pop-$\mathcal{B}$. 
The evaluation budget per generation is $\mathcal{O}((N_A+N_B)(R+1)K)$ fitness calls, where $K$ is the number of repeated labeled resamples per joint evaluation.

\noindent\textbf{Joint evaluation and scalar fitness.}
A joint pair $(a,b)$ is evaluated by executing the pseudo-labeling loop (Algorithm~\ref{alg:pseudolabel}) and scoring MacroF1 on the validation set (Eq.~\eqref{eq:macro_f1} in Section~\ref{sec:metrics}). Each evaluation repeats the SSL loop over $K$ labeled resamples with fixed seeds, producing scores $\{S_k(a,b)\}_{k=1}^{K}$ and summary statistics $\mu(a,b)$ and $\sigma(a,b)$. CC-SSL also logs auxiliary SSL diagnostic metrics defined in Section~\ref{sec:metrics}: pseudo-label volume (\texttt{pseudo\_added}, $\overline{n}_{\mathrm{add}}(a,b)$) and probe-drop (\texttt{probe\_drop}, $\mathrm{ProbeDrop}(a,b)$). The scalar fitness is defined in Eq.~\eqref{eq:fitness_ccssl}, where $\lambda_{\mathrm{std}}, \lambda_{\mathrm{bias}}, \lambda_{\mathrm{add}} \ge 0$ weights are tuned.

\begin{equation}
\label{eq:fitness_ccssl}
\footnotesize
F(a,b)=\mu(a,b)
-\lambda_{\mathrm{std}}\sigma(a,b)
-\lambda_{\mathrm{bias}}\,\mathrm{ProbeDrop}(a,b)
-\lambda_{\mathrm{add}}\,\overline{n}_{\mathrm{add}}(a,b),
\end{equation}

\noindent\textbf{Variation and repair operators.}
Each population uses variation operators tailored to its genotype structure. Operators are applied independently within each population.

Regarding Pop-$\mathcal{A}$, binary mask genes undergo bit-flip mutation with probability $p_{\mathrm{bit}}$. A repair operator enforces $\lVert m^{(v)}\rVert_0 \ge k_{\min}$ by activating randomly selected zero bits until the constraint holds. Discrete integer genes are perturbed by $\pm1$ steps and clamped to valid ranges. Boolean flags flip with probability $p_{\mathrm{flip}}$. Uniform crossover swaps individual genes independently with probability~$0.5$.

Regarding Pop-$\mathcal{B}$, 
continuous genes are perturbed by bounded additive noise and clamped to predefined intervals. Integer genes $(q,T)$ mutate by $\pm1$ steps and are clamped to valid ranges. Boolean gene $\nu$ flips with probability $p_{\mathrm{flip}}$. Arithmetic crossover is applied to continuous genes, while discrete genes are swapped with probability $0.5$. A feasibility repair enforces $\tau_{\min}\le\tau_0$ and $\Delta\tau\ge 0$.

\noindent\textbf{Selection and replacement.}
Each population evolves independently via tournament selection \cite{eiben2015ec}.
%Given tournament size $s$, selection draws $s$ individuals uniformly at random without replacement and returns the individual with the highest assigned fitness.
Selection is applied within each population only.
After offspring generation, CC-SSL applies generational replacement with elitism independently in each population.
The top $E$ individuals from the current population are copied to the next generation.
The remaining slots are filled by offspring in fitness-descending order (ties broken randomly).

\begin{algorithm}[t]
\caption{CC-SSL: CC with elitist+random collaboration}
\label{alg:ccssl_compact}
\begingroup
\renewcommand{\baselinestretch}{0.7}\selectfont
\begin{algorithmic}[1]
\Require $L_0,U_0$, $N_A,N_B,G$, $R$, $K$, EA operators $\Omega_{\mathcal{A}},\Omega_{\mathcal{B}}$
\State Initialize Pop-$\mathcal{A}$ and Pop-$\mathcal{B}$ with feasible genotypes
\State Set $(a^\star,b^\star)\leftarrow \arg\max F(a,b)$ over initial collaborations
\For{$g=1,\dots,G$}
  \State $a^{\mathrm{best}}\leftarrow \arg\max_{a\in \text{Pop-}\mathcal{A}}\mathrm{fit}(a)$
  \State $b^{\mathrm{best}}\leftarrow \arg\max_{b\in \text{Pop-}\mathcal{B}}\mathrm{fit}(b)$
  \ForAll{$a_i\in\text{Pop-}\mathcal{A}$}
    \State Sample $\{b^{(r)}\}_{r=1}^{R}\subset\text{Pop-}\mathcal{B}$ uniformly
    \State Set $\mathrm{fit}(a_i)\leftarrow \max_{b\in\mathcal{P}_B(a_i)} F(a_i,b)$ using $K$ resamples
  \EndFor
  \ForAll{$b_j\in\text{Pop-}\mathcal{B}$}
    \State Sample $\{a^{(r)}\}_{r=1}^{R}\subset\text{Pop-}\mathcal{A}$ uniformly
    \State Set $\mathrm{fit}(b_j)\leftarrow \max_{a\in\mathcal{P}_A(b_j)} F(a,b_j)$ using $K$ resamples
  \EndFor
  \State Update $(a^\star,b^\star)$ with the best joint-evaluated pair in $g$
  \State Pop-$\mathcal{A}\leftarrow \textsc{Evolve}(\text{Pop-}\mathcal{A}, \mathrm{fit},\Omega_{\mathcal{A}})$
  \State Pop-$\mathcal{B}\leftarrow \textsc{Evolve}(\text{Pop-}\mathcal{B}, \mathrm{fit},\Omega_{\mathcal{B}})$
\EndFor
\State \Return $(a^\star,b^\star)$
\end{algorithmic}
\endgroup
\end{algorithm}

\noindent\textbf{Monolithic EA-SSL.}
EA-SSL applies a standard EA which uses the same tournament selection, variation and repair operators, elitist replacement scheme, scalar fitness function, and evaluation budget as CC-SSL. This design isolates the effect of problem factorization and collaborative credit assignment (two-population cooperative coevolution) versus monolithic encoding (single-population EA) under shared building blocks and matched representation capacity. Due to space constraints, the full EA-SSL procedure is not detailed; it follows directly from Algorithm~\ref{alg:ccssl_compact} by removing partner sampling and evolving a single population.

% =========================
% Experimental Setup
% =========================
\section{Experimental Setup}
\label{sec:exp_settings}

\sloppy
We empirically evaluate CC-SSL and EA-SSL on a heterogeneous tabular benchmark from OpenML~\cite{vanschoren2013openml}. The study uses $25$ datasets and separates a \emph{tuning set} from a \emph{benchmark set}. 
Tuning relies on $\mathcal{D}_{\mathrm{tune}}=\{28,44,46\}$, while benchmarking uses $\mathcal{D}_{\mathrm{bench}}$, comprising $11$ binary datasets $\{3,15,29,31,37,38,44,50,1049,1067,1480\}$~and~$11$ multiclass datasets $\{11,12,14,16,22,23,28,46,54,307,1468\}$. The labeled set size is determined by labeled fractions $\mathcal{L}=\{0.01,0.05,0.10\}$ relative to the SSL training pool (1\%, 5\%, and 10\% labeled samples, respectively). The datasets span $625$ to $5620$ rows and $5$ to $857$ features. Table~\ref{tab:datasets} in the Appendix reports dataset characteristics.

%We empirically evaluate CC-SSL and EA-SSL on a heterogeneous tabular benchmark obtained from OpenML~\cite{vanschoren2013openml}. The study uses $25$ datasets and separates a \emph{tuning set} from a \emph{benchmark set} for final evaluation. Tuning relises on $\mathcal{D}_{\mathrm{tune}}=\{28,44,46\}$, while benchmarking uses $\mathcal{D}_{\mathrm{bench}}$, comprising $11$ binary datasets $\{3,15,29,31,37,38,44,50,1049,1067,1480\}$ and $11$ multiclass datasets $\{11,12,14,16,22,23,28,46,54,1468,1475,307\}$. The labeled set size is determined by labeled fractions $\mathcal{L}=\{0.01,0.05,0.10\}$ relative to the SSL training pool (1\%, 5\%, and 10\% labeled samples, respectively). The datasets span $625$ to $5620$ rows and $5$ to $857$ features after preprocessing. Table~\ref{tab:datasets} in the Appendix reports dataset characteristics.

The evaluation includes three classical lightweight SSL baselines: \textbf{Self-training} (\texttt{ST}, pseudo-labeling with a fixed policy), \textbf{Label Spreading} (\texttt{LS}, graph-based label propagation/spreading), and \textbf{Heuristic co-training} (\texttt{HCo}, two random feature splits with a fixed policy). Baselines use the same data partitions and preprocessing as the evolutionary methods.
%A supervised-only reference (train only on $L$) can be reported as context when interpreting low-label results.

All methods were implemented in Python 3, using NumPy and scikit-learn libraries.
The code is publicly available at GitHub~\cite{ccssl_github}. Computational experiments were executed on a high-performance cluster consisting of 126 SD530 nodes, each equipped with 52 Intel Xeon Gold 6230R cores operating at 2.10 GHz and 192 GB of RAM. 

\subsection{Evaluated metrics}
\label{sec:metrics}
\vspace{-0.1cm}
This paper reports MacroF1 (MacroF1), classification accuracy (Acc), and population diversity.
The experimental evaluation also considers pseudo-label volume, probe-drop, and validation optimism to characterize pseudo-label dynamics and generalization behavior, and wall-clock time as a computational cost proxy.

This paper reports MacroF1 (MacroF1), classification accuracy (Acc), and population diversity.
The experimental evaluation also considers pseudo-label volume, probe-drop, and validation optimism to characterize pseudo-label dynamics and generalization behavior, and wall-clock time as a computational cost proxy.

\vspace{-0.1cm}
\paragraph{Predictive performance metrics.}
Let $\mathcal{T}=\{(x_i,y_i)\}_{i=1}^{n_{\text{test}}}$ be the labeled test set with $C$ classes, and let $\hat{y}_i$ denote the predicted class for $x_i$.
For each class $c\in\{1,\dots,C\}$, define one-vs-rest counts on $\mathcal{T}$: true positives $TP_c$, false positives $FP_c$, and false negatives $FN_c$.

For each class $c$, precision $\mathrm{Prec}_c$, recall $\mathrm{Rec}_c$, and class-wise F1-score $\mathrm{F1}_c$ are defined in Eq.~\eqref{eq:f1}. MacroF1 averages $\mathrm{F1}_c$ uniformly across classes (Eq.~\eqref{eq:macro_f1}). MacroF1 assigns equal weight to each class, which is appropriate for mixed binary/multiclass benchmarks and class-imbalance settings~\cite{sokolova2009systematic}.
\begin{equation}
\label{eq:f1}
\mathrm{Prec}_c=\frac{TP_c}{TP_c+FP_c},\ 
\mathrm{Rec}_c=\frac{TP_c}{TP_c+FN_c},\ 
\mathrm{F1}_c=\frac{2\,\mathrm{Prec}_c\,\mathrm{Rec}_c}{\mathrm{Prec}_c+\mathrm{Rec}_c}.
\vspace{-0.1cm}
\end{equation}
\begin{equation}
\label{eq:macro_f1}
\mathrm{MacroF1}(\mathcal{T})=\frac{1}{C}\sum_{c=1}^{C}\mathrm{F1}_c.
\end{equation}

\paragraph{Pseudo-label dynamics and generalization diagnostics}
%\paragraph{Pseudo-label volume (aggressiveness proxy).} 
%Pseudo-label volume is measured as aggressiveness proxy. 
Let $L_0$ denote the initial labeled set and let $L_T$ denote the labeled set after $T$ pseudo-labeling iterations (Algorithm~\ref{alg:pseudolabel}).
Pseudo-label volume is the number of accepted pseudo-labels ($\overline{n}_{\mathrm{add}}$), where $\Delta L_t$ is the set of unlabeled samples accepted at iteration $t$ (Eq~\eqref{eq:pseudo_added}).
Larger \texttt{pseudo\_added} corresponds to a more aggressive acceptance policy in self-training and pseudo-labeling procedures~\cite{triguero2015selflabeled,vanengelenhoos2020survey}. 

\begin{equation}
\label{eq:pseudo_added}
\overline{n}_{\mathrm{add}}=|L_T|-|L_0|=\sum_{t=1}^{T}|\Delta L_t|,
\end{equation}

%\paragraph{Probe-drop (confirmation-bias proxy).}
%Probe-drop is evaluated as a confirmation-bias proxy.
Let $P$ be a held-out labeled \emph{probe} set that is never augmented with pseudo-labels.
Let $S(\cdot)$ denote the evaluation metric (e.g., MacroF1).
Let $S_P^{(0)}$ be the score on $P$ for a model trained on $L_0$, and let $S_P^{(T)}$ be the score on $P$ after completing the pseudo-labeling loop (trained on $L_T$).
Probe-drop is defined as: $\mathrm{ProbeDrop}=S_P^{(0)}-S_P^{(T)}$.
Positive values indicate degradation on $P$ after adding pseudo-labels, which is consistent with confirmation bias in self-labeled SSL~\cite{triguero2015selflabeled,vanengelenhoos2020survey}.

%\paragraph{Validation optimism (generalization-gap proxy).}
Let $V$ be the validation set.
Let $S_V$ be the score on $V$ and let $S_{\mathcal{T}}$ be the score on the test set $\mathcal{T}$ under the same training procedure.
Validation optimism is defined as: $\texttt{val\_minus\_test}=S_V - S_{\mathcal{T}}$.
Positive values indicate optimistic validation estimates.  relative to test performance.

\paragraph{Population diversity and computational cost.}
Population diversity is computed using metrics matched to the genotype representation.
For the view population (Pop-$\mathcal{A}$), which encodes boolean feature masks, diversity is the average pairwise Jaccard distance.
For the policy population (Pop-$\mathcal{B}$), which encodes heterogeneous parameters, diversity is computed separately for numeric and boolean components.
Continuous and integer parameters (e.g., $\tau_{start}$, $max_{iters}$) are min--max normalized to $[0,1]$ using their domain bounds, and numeric diversity is the mean Euclidean distance to the centroid.
Boolean parameters (e.g., \texttt{calibrate}) are summarized by the mean disagreement rate, and boolean diversity $D_{\text{bool}}$ is the mean disagreement rate averaged across boolean flags.

%\paragraph{Cost analysis (runtime and convergence proxies).}
Computational cost is analyzed using two convergence-oriented proxies: \emph{Time-to-Target} (TTT) and \emph{Generations-to-Target} (GTT).
Let $F^{\star}$ denote the final best-so-far fitness attained in a run.
The target fitness is defined as $0.99\,F^{\star}$.
Generations-to-Target, $\texttt{GTT}_{0.99}$, is the first generation index at which the best-so-far fitness reaches or exceeds $0.99\,F^{\star}$.
Time-to-Target, $\texttt{TTT}_{0.99}$, is the corresponding cumulative wall-clock time elapsed up to that generation.
TTT captures practical runtime efficiency, while GTT isolates algorithmic convergence behavior independent of per-generation cost.
%Both metrics are computed per run and summarized across seeds and datasets using robust statistics.

\subsection{Parameter settings and sensitivity analysis}
\label{sec:tuning_protocol}

\begin{figure}[!h]
\setlength{\abovecaptionskip}{-6pt}
\setlength{\belowcaptionskip}{-6pt}
    \centering
    \includegraphics[width=0.82\linewidth]{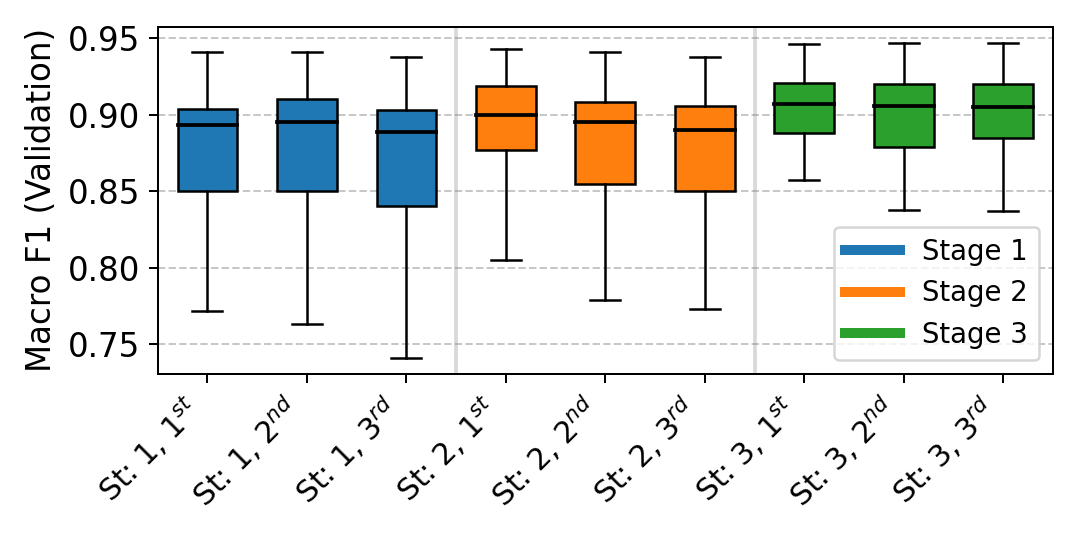}
    \caption{Parameter tuning results summary}
    \label{fig:tuning_boxplot}
\end{figure}

The CC-SSL configuration involves $10$ hyperparameters grouped into population sizes and budgets ($3$ parameters), genetic operator probabilities ($4$ parameters), and scalar fitness weights ($3$ parameters), each with multiple plausible candidate values (Table~\ref{tab:param_settings_staged} in Appendix). An exhaustive joint search over this space is statistically inefficient given stochastic variance in evolutionary SSL. Therefore, we adopt a three-stage tuning protocol that decomposes the configuration into progressively finer decisions, following established practice in evolutionary computation and experimental design to reduce dimensionality, control confounding effects, and limit overfitting to the tuning data~\cite{eiben2007parameter}. All tuning is restricted to the development set $\mathcal{D}_{\mathrm{tune}}=\{28,44,46\}$ and labeled fractions $\mathcal{L}=\{0.01,0.05,0.10\}$. 
%After tuning, the selected parameters are frozen and used unchanged for the benchmark evaluation on $\mathcal{D}_{\mathrm{bench}}$. 
Each configuration is run 15~times. Figure~\ref{fig:tuning_boxplot} summarizes parameter tuning results by showing the distribution of results of the three best configurations per stage.

\paragraph{Stage 1: population sizes and generations (cost-matched budgets).}
Stage 1 compares three population-size configurations while approximately matching total compute by adjusting the number of generations.
The number of collaborators per evaluated individual is fixed to \texttt{teams\_per\_individual}$=3$.
The evaluated budgets are:
\[
( |popA|,|popB|,G )\in\{(6,6,50),\ (8,8,38),\ (10,10,30)\}.
\]
The rationale is to keep the total number of team evaluations of the same order:
\[
\text{Total team evals} \approx G \cdot \texttt{teams\_per\_individual} \cdot (|A|+|B|),
\]
while varying the depth--breadth trade-off (more generations vs.\ larger populations).
Selection uses mean test MacroF1 on $\mathcal{D}_{\mathrm{tune}}$ aggregated across labeled fractions as the primary criterion.

The evaluated Stage~1 configurations did not show statistically significant differences in final performance. Therefore, we compared practical convergence using time-to-target at 99\% of the final best fitness, denoted $TTT_{0.99}$. The $TTT_{0.99}$ results show diminishing returns across configurations. Runs with smaller populations and longer budgets $(|popA|,|popB|,G)=(6,6,50)$ reached stable plateau behavior more consistently than larger populations under cost-matched generation budgets. This pattern indicates that additional selection--variation cycles can improve stabilization under cooperative credit assignment.

\paragraph{Stage 2: evolutionary operator probabilities.}
Stage~2 tunes crossover and mutation probabilities for each population using a full-factorial grid with $3^4=81$ operator configurations. The primary criterion is mean test MacroF1 aggregated over $\mathcal{D}_{\mathrm{tune}}\times\mathcal{L}$ and seeds. The top configurations do not show statistically significant differences under this criterion. Therefore, selection uses lower MacroF1 dispersion (standard deviation) across seeds as a tie-breaker. The selected configuration is $p^{(A)}_{cx}=0.85$, $p^{(A)}_{mut}=0.45$, $p^{(B)}_{cx}=0.85$, and $p^{(B)}_{mut}=0.35$.

\paragraph{Stage 3: scalar-fitness weights.}
Stage~3 tunes the weights of the scalar fitness in Eq.~\eqref{eq:fitness_ccssl}. A grid of $27$ candidate weight configurations is evaluated on the development blocks induced by $\mathcal{D}_{\mathrm{tune}}\times\mathcal{L}$ and the tuning seeds. The primary criterion is mean test MacroF1 aggregated. When configurations are not statistically distinguishable under this criterion, selection uses lower MacroF1 dispersion across seeds as a tie-breaker. The selected configuration is $\lambda_{std}=0.4$, $\lambda_{bias}=0.7$, and $\lambda_{added}=0.0$. Setting $\lambda_{added}=0.0$ removes the pseudo-label volume penalty term from the fitness.

\textit{EA-SSL parameter settings.}
EA-SSL is configured to match the CC-SSL search budget and objective. EA-SSL uses $G=50$ and a population size of $36$ individuals to keep the total number of evaluated solutions of the same order as CC-SSL under the team-evaluation protocol. EA-SSL uses the same scalar fitness weights as CC-SSL to preserve the optimization objective. Operator probabilities are tuned using the Stage~2 protocol. The selected configuration is $p^{(A)}_{cx}=0.85$, $p^{(A)}_{mut}=0.35$, $p^{(B)}_{cx}=0.85$, and $p^{(B)}_{mut}=0.35$.

\section{Experimental analysis}
\label{sec:experimental_results}

This section evaluates CC-SSL and EA-SSL against lightweight SSL baselines and standard supervised classifiers under a controlled and statistically rigorous protocol. Parameter tuning is performed using $15$ independent runs and benchmark results are obtained from~$30$~independent runs. Statistical comparisons use Wilcoxon signed-rank tests with significance level $\alpha=0.01$, applying Bonferroni correction when required. Results are analyzed separately for binary and multiclass datasets, as these settings differ substantially in class structure, decision boundaries, and sensitivity to pseudo-label noise. Comparisons include three SSL baselines: Self-Training (\texttt{ST}), Heuristic Co-Training (\texttt{HCo}), and Label Spreading (\texttt{LS}), as well as regular supervised classifiers (Logistic Regression, Linear Support Vector Machine, Random Forest, and Histogram-based Gradient Boosting Classification Tree) to contextualize absolute performance. The analysis proceeds from final test performance, to evolutionary dynamics and cost, and finally to pseudo-label behavior and stability.

\subsection{Final results comparison}\label{subsec:final_results}
This subsection reports final test performance for CC-SSL and EA-SSL against three lightweight SSL baselines.
The primary metric is test MacroF1; test accuracy is secondary.
Tables~\ref{tab:final:f1-bin} and~\ref{tab:final:f1-multi} summarize MacroF1 for binary and multiclass datasets, respectively.
Accuracy summaries and full data-wise results are reported in the appendix (Tables~\ref{tab:final:acc-binary}--\ref{tab:full:acc-multiclass}).
The rows \textit{Best Class.} and \textit{Global Class.} report purely supervised reference results: \textit{Best Class.} is the best-performing classifier, and \textit{Global Class.} aggregates results across all evaluated classifiers.
These supervised references provide a strong non-SSL comparator, while the evolutionary SSL pipeline uses a lightweight linear base learner (logistic regression) in the pseudo-labeling loop.

An evolutionary method \emph{wins} on a dataset at labeled fraction $lf$ when it is statistically better than \emph{each} SSL baseline under a paired Wilcoxon signed-rank test across seeds with Bonferroni correction ($\alpha=0.01$).
Win counts in Tables~\ref{tab:final:f1-bin}--\ref{tab:final:f1-multi} summarize this criterion.

\begin{table}[h]
\renewcommand{\arraystretch}{0.95}
\centering
\small
\caption{Descriptive statistics for MacroF1 results for 11~binary datasets} \label{tab:final:f1-bin}
\begin{tabular}{llrrrrr}
\toprule
$lf$ & Method & Min & Median & IQR & Max & Wins \\
\midrule
0.01 & CC-SSL & 0.540 & 0.675 & 0.214 & 0.948 & 7 \\
0.01 & EA-SSL & 0.579 & 0.694 & 0.235 & 0.953 & 7 \\
0.01 & \texttt{ST} & 0.357 & 0.484 & 0.167 & 0.936 & - \\
0.01 & \texttt{LS} & 0.343 & 0.421 & 0.090 & 0.760 & - \\
0.01 & \texttt{HCo} & 0.343 & 0.496 & 0.185 & 0.912 & - \\
0.01 & Best Class. & 0.522 & 0.632 & 0.146 & 0.948 & - \\
0.01 & Global Class. & 0.497 & 0.578 & 0.135 & 0.916 & - \\
\hline
0.05 & CC-SSL & 0.593 & 0.696 & 0.236 & 0.943 & 8 \\
0.05 & EA-SSL & 0.571 & 0.711 & 0.256 & 0.953 & 8 \\
0.05 & \texttt{ST} & 0.437 & 0.600 & 0.221 & 0.936 & - \\
0.05 & \texttt{LS} & 0.308 & 0.412 & 0.090 & 0.694 & - \\
0.05 & \texttt{HCo} & 0.453 & 0.543 & 0.194 & 0.953 & - \\
0.05 & Best Class. & 0.585 & 0.692 & 0.253 & 0.957 & - \\
0.05 & Global Class. & 0.548 & 0.665 & 0.238 & 0.945 & - \\
\hline
0.10 & CC-SSL & 0.605 & 0.688 & 0.235 & 0.942 & 6 \\
0.10 & EA-SSL & 0.577 & 0.697 & 0.236 & 0.948 & 4 \\
0.10 & \texttt{ST} & 0.495 & 0.637 & 0.263 & 0.935 & - \\
0.10 & \texttt{LS} & 0.343 & 0.403 & 0.072 & 0.608 & - \\
0.10 & \texttt{HCo} & 0.422 & 0.565 & 0.312 & 0.942 & - \\
0.10 & Best Class. & 0.597 & 0.799 & 0.220 & 0.953 & - \\
0.10 & Global Class. & 0.553 & 0.676 & 0.215 & 0.938 & - \\
\bottomrule
\end{tabular}
\end{table}

\begin{table}[!h]
\renewcommand{\arraystretch}{0.95}
\centering
\small
\caption{Descriptive statistics for MacroF1 results for 11~multiclass datasets} \label{tab:final:f1-multi}
\begin{tabular}{llrrrrr}
\toprule
$lf$ & Method & Min & Median & IQR & Max & Wins \\
\midrule
0.01 & CC-SSL & 0.368 & 0.615 & 0.255 & 0.872 & 11 \\
0.01 & EA-SSL & 0.387 & 0.628 & 0.275 & 0.913 & 11 \\
0.01 & \texttt{ST} & 0.313 & 0.399 & 0.242 & 0.709 & - \\
0.01 & \texttt{LS} & 0.022 & 0.171 & 0.181 & 0.383 & - \\
0.01 & \texttt{HCo} & 0.295 & 0.463 & 0.146 & 0.661 & - \\
0.01 & Best Class. & 0.320 & 0.558 & 0.205 & 0.777 & - \\
0.01 & Global Class. & 0.292 & 0.475 & 0.159 & 0.726 & - \\
\midrule
0.05 & CC-SSL & 0.358 & 0.716 & 0.310 & 0.923 & 8 \\
0.05 & EA-SSL & 0.354 & 0.696 & 0.296 & 0.934 & 10 \\
0.05 & \texttt{ST} & 0.320 & 0.584 & 0.340 & 0.904 & - \\
0.05 & \texttt{LS} & 0.015 & 0.043 & 0.160 & 0.228 & - \\
0.05 & \texttt{HCo} & 0.279 & 0.606 & 0.248 & 0.897 & - \\
0.05 & Best Class. & 0.352 & 0.669 & 0.233 & 0.912 & - \\
0.05 & Global Class. & 0.292 & 0.608 & 0.225 & 0.880 & - \\
\midrule
0.10 & CC-SSL & 0.395 & 0.754 & 0.281 & 0.942 & 6 \\
0.10 & EA-SSL & 0.413 & 0.747 & 0.274 & 0.943 & 6 \\
0.10 & \texttt{ST} & 0.344 & 0.709 & 0.296 & 0.935 & - \\
0.10 & \texttt{LS} & 0.015 & 0.078 & 0.166 & 0.228 & - \\
0.10 & \texttt{HCo} & 0.303 & 0.721 & 0.272 & 0.923 & - \\
0.10 & Best Class. & 0.441 & 0.730 & 0.248 & 0.931 & - \\
0.10 & Global Class. & 0.373 & 0.699 & 0.245 & 0.923 & - \\
\bottomrule
\end{tabular}
\end{table}

\paragraph{Binary datasets}
CC-SSL and EA-SSL achieve higher median MacroF1 than \texttt{ST}, \texttt{LS}, and \texttt{HCo} at all labeled fractions (Table~\ref{tab:final:f1-bin}).
The separation is largest at $lf=0.01$, and the weakest results are observed for \texttt{LS}.
At $lf=0.10$, \textit{Best Class.} is competitive and attains a higher median than both evolutionary methods, indicating that a strong supervised classifier can match or exceed the SSL pipeline when the labeled pool is larger.
Accuracy follows the same qualitative ordering (Appendix Table~\ref{tab:final:acc-binary}).

\paragraph{Multiclass datasets}
On multiclass datasets (Table~\ref{tab:final:f1-multi}), CC-SSL and EA-SSL attain higher medians than the three SSL baselines at all labeled fractions, with the largest separation at $lf=0.01$.
Relative to supervised references, both evolutionary methods are higher at $lf=0.01$, while the gaps narrow at $lf\in\{0.05,0.10\}$.
In the same setting, \texttt{ST} becomes more competitive as $lf$ increases and its median approaches the evolutionary methods.
Accuracy results are consistent and are reported in the appendix (Table~\ref{tab:final:acc-multiclass}).

\paragraph{CC-SSL vs.\ EA-SSL}
At $lf=0.01$ (i.e., the most critical regime), CC-SSL and EA-SSL achieve substantially higher median MacroF1 than SSL baselines on both binary and multiclass datasets (Tables~\ref{tab:final:f1-bin}--\ref{tab:final:f1-multi}), with the largest separation observed against \texttt{LS}. 
As $lf$ increases, \texttt{ST} becomes more competitive and the median gaps narrow.
Focusing on CC-SSL and EA-SSL, across all runs, CC-SSL attains higher MacroF1 and accuracy in 38\% of cases, while EA-SSL attains higher values in 62\%.
Under paired Wilcoxon tests across seeds, most dataset--$lf$ comparisons are draws for both MacroF1 and accuracy.
This outcome indicates limited evidence of systematic final-performance differences between CC-SSL and EA-SSL under the matched representation budget, despite differences observed in evolutionary trajectories.

%The direct CC-SSL vs.\ EA-SSL comparison again produces mostly draws. For MacroF1, EA-SSL wins once at $lf=0.01$ and all remaining cases are draws. For accuracy, EA-SSL wins once at $lf=0.01$ and once at $lf=0.05$. This pattern supports the same conclusion as in the binary setting.

%\paragraph{RQ0 (Context): Do evolutionary SSL methods outperform lightweight SSL baselines in the extreme low-label regime?}Yes. At $lf=0.01$, CC-SSL and EA-SSL achieve nonzero win counts against all three baselines on binary datasets ($5/8$ each) and win on all multiclass datasets ($9/9$ each) in test MacroF1. Under the same criterion, all baselines have $0$ wins in both binary and multiclass settings (Tables~\ref{tab:final:f1-bin} and~\ref{tab:final:f1-multi}).

\textit{\textbf{RQ1:} Do evolutionary SSL methods outperform lightweight SSL baselines in the extreme low-label regime?}
Yes. At $lf=0.01$, CC-SSL and EA-SSL obtain nonzero win counts against \texttt{ST}, \texttt{LS}, and \texttt{HCo} on binary datasets (Table~\ref{tab:final:f1-bin}) and win on all multiclass datasets (Table~\ref{tab:final:f1-multi}) in test MacroF1. Under the same criterion, the three SSL baselines obtain zero wins in both binary and multiclass settings.

\textit{\textbf{RQ2:} When does CC improve SSL relative to monolithic evolutionary search under matched budgets?} The benchmark does not show a systematic final-test advantage for CC-SSL over EA-SSL. Most dataset--$lf$ paired comparisons are draws for MacroF1 and accuracy under paired Wilcoxon tests across seeds ($\alpha=0.01$). EA-SSL attains a small number of isolated wins, while CC-SSL attains none in these direct comparisons.  These outcomes indicate similar final predictive performance for CC-SSL and EA-SSL. The subsequent analysis reports evolutionary trajectories and cost proxies to characterize differences beyond final test scores.

%\paragraph{RQ1 (Main): When does cooperative coevolution improve SSL relative to monolithic evolutionary search under the same representation budget?} The benchmark results do not show a systematic advantage of cooperative coevolution in final test performance. CC-SSL and EA-SSL draw in the majority of dataset-level paired comparisons for both MacroF1 and accuracy at all labeled fractions. EA-SSL attains a small number of isolated wins (MacroF1: one win at $lf=0.01$ in binary and one win at $lf=0.01$ in multiclass; accuracy: one win at $lf=0.01$ in binary and two wins at $lf\in\{0.01,0.05\}$ in multiclass), while CC-SSL attains no wins in these direct comparisons. These outcomes indicate similar final predictive performance for CC-SSL and EA-SSL under the matched search representation, motivating a closer analysis of convergence behavior and computational efficiency in subsequent sections.

\subsection{Evolutionary dynamics and computation cost-to-target analysis}\label{subsec:evo_dynamics}

\begin{figure}[!h]
\setlength{\abovecaptionskip}{2pt}
\setlength{\belowcaptionskip}{0pt}
\centering
\begin{subfigure}[t]{0.48\linewidth}
\setlength{\abovecaptionskip}{0pt}

  \centering
  \includegraphics[width=0.85\linewidth]{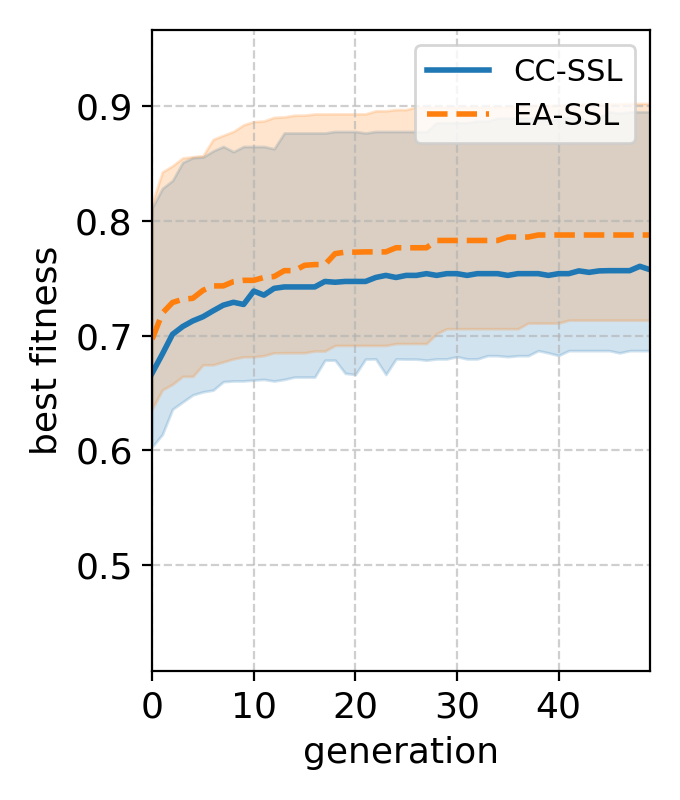}
  \caption{Binary datasets, $lf$=0.01}
  \label{fig:fitness_evo_bin_001}
\end{subfigure}\hfill
\begin{subfigure}[t]{0.48\linewidth}
\setlength{\abovecaptionskip}{0pt}

  \centering
  \includegraphics[width=0.85\linewidth]{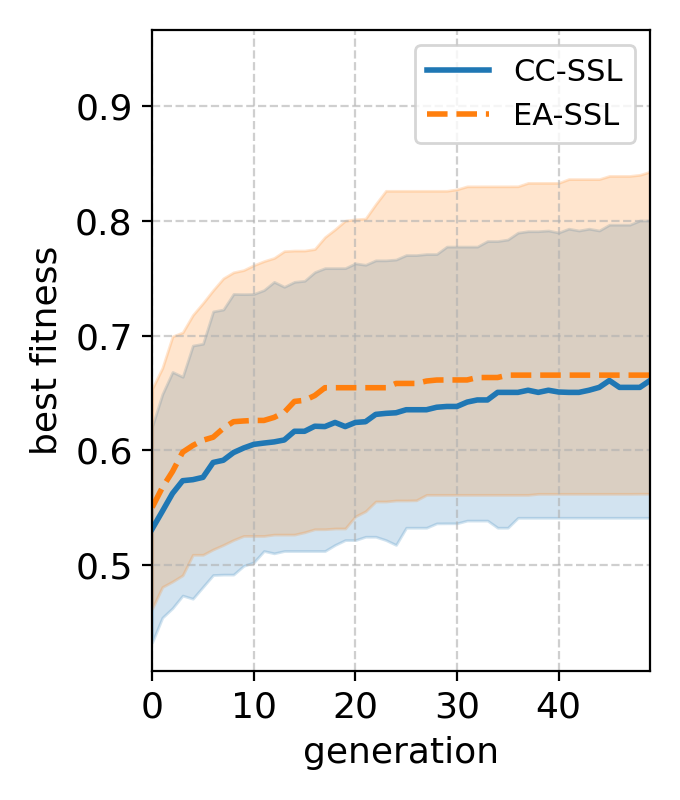}
  \caption{Multiclass datasets, $lf$=0.01}
  \label{fig:fitness_evo_multiclass_001}
\end{subfigure}

\begin{subfigure}[t]{0.48\linewidth}
\setlength{\abovecaptionskip}{0pt}

  \centering
  \includegraphics[width=0.85\linewidth]{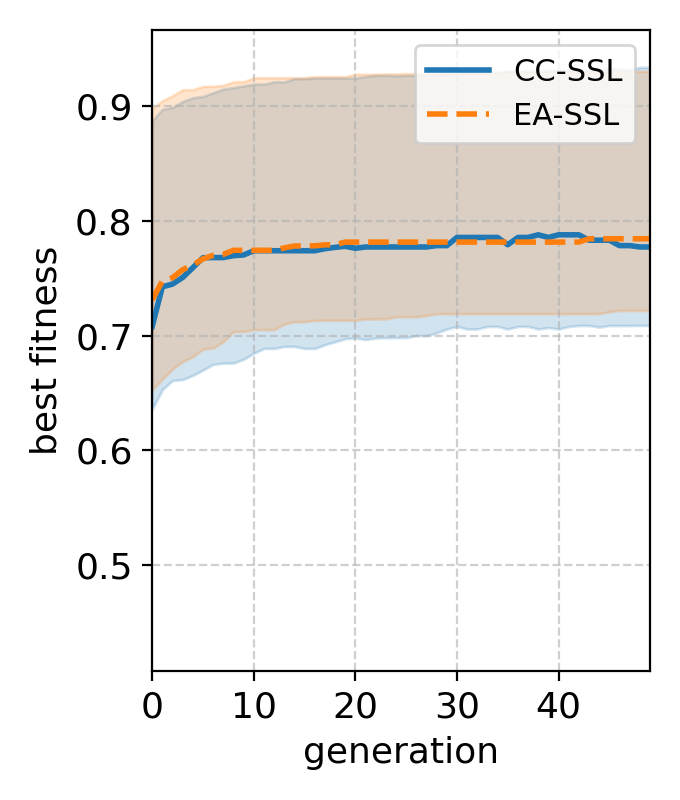}
  \caption{Binary datasets, $lf$=0.10}
  \label{fig:fitness_evo_bin_010}
\end{subfigure}\hfill
\begin{subfigure}[t]{0.48\linewidth}
\setlength{\abovecaptionskip}{0pt}

  \centering
  \includegraphics[width=0.85\linewidth]{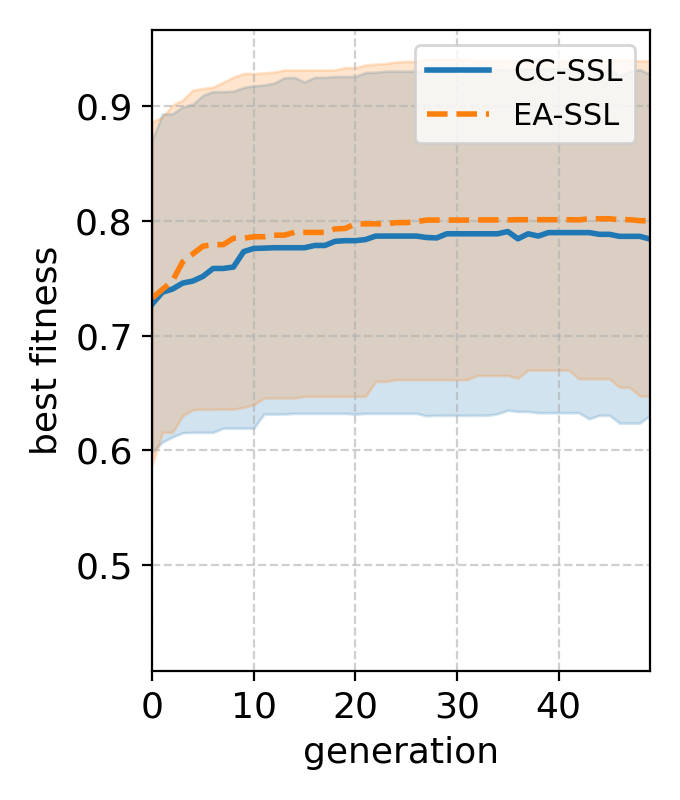}
  \caption{Multiclass datasets, $lf$=0.10}
  \label{fig:fitness_evo_multiclass_010}
\end{subfigure}
\caption{Median best-so-far fitness per generation}
\label{fig:fitness_evo}
\end{figure}

\begin{figure}[!h]
\setlength{\abovecaptionskip}{2pt}
\setlength{\belowcaptionskip}{-2pt}
\centering
\begin{subfigure}[t]{0.48\linewidth}
\setlength{\abovecaptionskip}{0pt}

  \centering
  \includegraphics[width=0.85\linewidth]{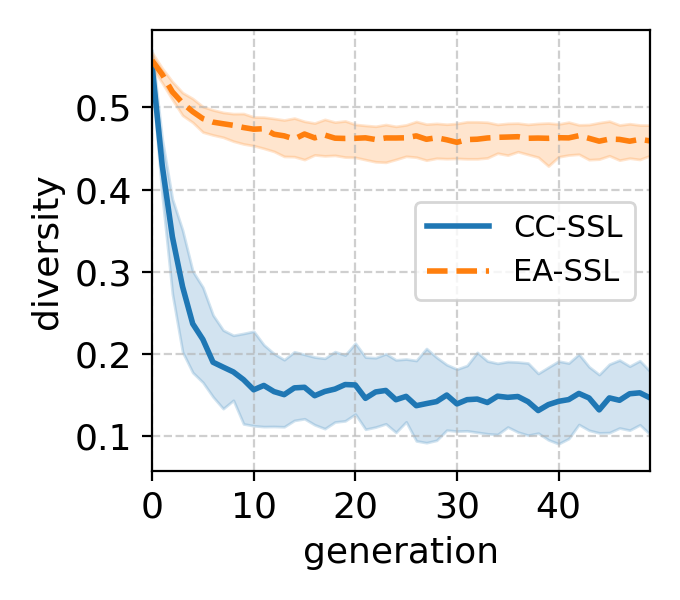}
  \caption{Binary datasets, $lf$=0.01}
  \label{fig:fitness_div_bin_001}
\end{subfigure}\hfill
\begin{subfigure}[t]{0.48\linewidth}
\setlength{\abovecaptionskip}{0pt}

  \centering
  \includegraphics[width=0.85\linewidth]{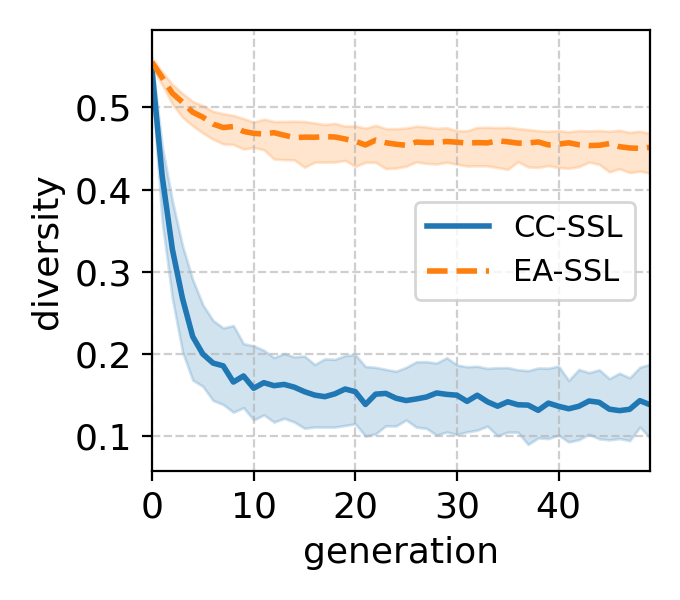}
  \caption{Multiclass datasets, $lf$=0.01}
  \label{fig:fitness_div_multiclass_001}
\end{subfigure}

\begin{subfigure}[t]{0.48\linewidth}
\setlength{\abovecaptionskip}{0pt}

  \centering
  \includegraphics[width=0.85\linewidth]{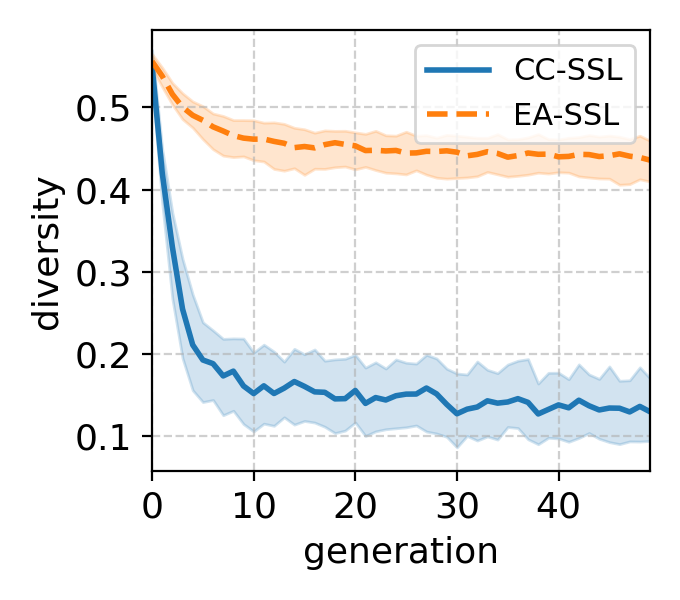}
  \caption{Binary datasets, $lf$=0.10}
  \label{fig:fitness_div_bin_010}
\end{subfigure}\hfill
\begin{subfigure}[t]{0.48\linewidth}
\setlength{\abovecaptionskip}{0pt}

  \centering
  \includegraphics[width=0.85\linewidth]{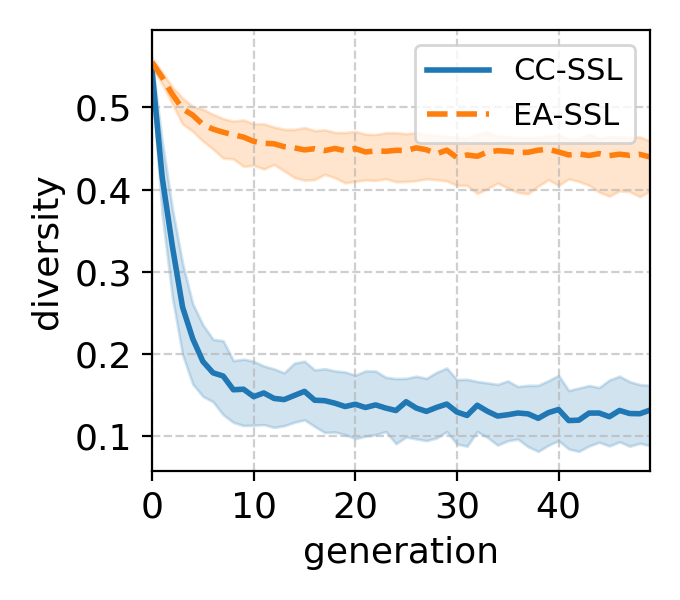}
  \caption{Multiclass datasets, $lf$=0.10}
  \label{fig:fitness_div_multiclass_010}
\end{subfigure}
\caption{Median population diversity per generation, measured as mean pairwise distance}
\label{fig:fitness_div}
\end{figure}

This subsection analyzes evolutionary dynamics of CC-SSL and \hbox{EA-SSL} reporting (i) best-so-far fitness trajectories across generations, (ii) population diversity as a proxy for exploration, and (iii) convergence cost to a per-run target fitness (i.e., TTT and GTT).
Figures~\ref{fig:fitness_evo} and~\ref{fig:fitness_div} report median trajectories aggregated, separately for binary and multiclass benchmarks, and for representative labeled fractions $lf\in\{0.01,0.10\}$ (with additional labeled fractions summarized in the TTT/GTT boxplots).

Figure~\ref{fig:fitness_evo} reports the median best-so-far fitness across generations. EA-SSL gets higher median fitness than CC-SSL at all labeled fractions, with the separation most pronounced at $lf=0.01$ and reduced as $lf$ increases. The gap appears in early generations and persists until termination. Final test performance differences between EA-SSL and CC-SSL are small and rarely statistically significant (Section~\ref{subsec:final_results}), so higher in-run fitness does not translate into systematic gains on held-out test performance in this benchmark.

Figure~\ref{fig:fitness_div} illustrates median population diversity over generations, measured as mean pairwise distance (Jaccard for view masks and Euclidean distance for numeric policy parameters). EA-SSL maintains higher diversity than CC-SSL at all labeled fractions. CC-SSL exhibits a faster early decrease in diversity, followed by a lower steady-state level. This gap is consistent with the population structures and the diversity metric: EA-SSL evolves a single population of $36$ individuals, whereas CC-SSL evolves two subpopulations of size $6$ each and reports diversity within each representation type. Under the same variation operators, larger populations increase the expected number of distinct pairs, which raises mean pairwise distance. In addition, CC-SSL assigns fitness via maximization over a limited collaborator set, which favors mutually compatible components and can reduce within-subpopulation dispersion.

%Figure~\ref{fig:fitness_div} illustrates median population diversity over generations, measured as mean pairwise distance (Jaccard for view masks and Euclidean distance for numeric policy parameters). EA-SSL maintains higher diversity than CC-SSL at all labeled fractions. CC-SSL exhibits a faster early decrease in diversity, followed by a lower steady-state level. This gap is consistent with the population structures and the diversity metric: EA-SSL evolves a single population of $36$ complete genotypes $(a,b)$, whereas CC-SSL evolves two subpopulations of size $6$ each and reports diversity within each representation type (mask Jaccard for Pop-$\mathcal{A}$ and numeric-policy distance for Pop-$\mathcal{B}$). Under the same variation operators, larger populations increase the expected number of distinct pairs, which raises mean pairwise distance. In addition, CC-SSL assigns fitness via maximization over a limited collaborator set, which favors mutually compatible components and can reduce within-subpopulation dispersion.

\begin{figure}[!h]
\setlength{\abovecaptionskip}{2pt}
\setlength{\belowcaptionskip}{-2pt}
\centering
\begin{subfigure}[t]{0.48\linewidth}
\setlength{\abovecaptionskip}{0pt}
 \centering
  \includegraphics[width=0.85\linewidth]{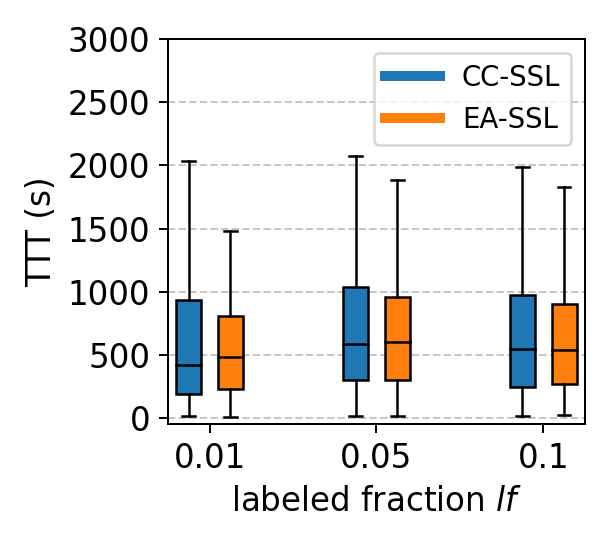}
  \caption{Binary datasets}
  \label{fig:ttt_boxplot_binary}
\end{subfigure}\hfill
\begin{subfigure}[t]{0.48\linewidth}
\setlength{\abovecaptionskip}{0pt}
  \centering
  \includegraphics[width=0.85\linewidth]{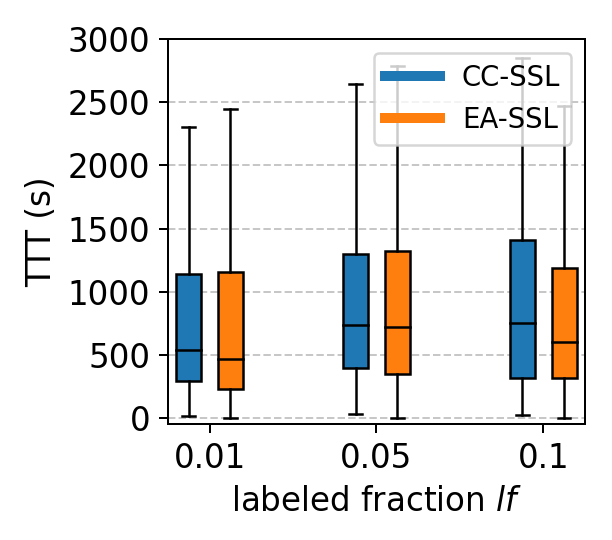}
  \caption{Multiclass datasets}
  \label{fig:ttt_boxplot_multiclass}
\end{subfigure}
\caption{Result distributions of TTT.}
\label{fig:ttt_dist}
\end{figure}

Figures~\ref{fig:ttt_boxplot_binary} and~\ref{fig:ttt_boxplot_multiclass} show the distributions of TTT for binary and multiclass datasets, respectively.
Across labeled fractions, median TTT values for CC-SSL and EA-SSL are comparable.
On binary datasets, no statistically significant differences are observed at any labeled fraction.
On multiclass datasets, EA-SSL shows slightly lower median TTT values, but differences are not statistically significant at individual labeled fractions.
Overall, TTT results indicate similar wall-clock time-to-quality for CC-SSL and EA-SSL.

\begin{figure}[!h]
\setlength{\abovecaptionskip}{2pt}
\setlength{\belowcaptionskip}{-2pt}
\centering
\begin{subfigure}[t]{0.48\linewidth}
\setlength{\abovecaptionskip}{0pt}
  \centering
  \includegraphics[width=0.92\linewidth]{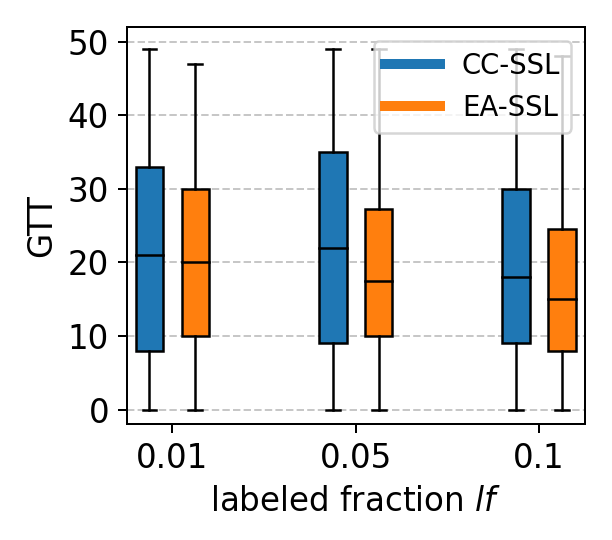}
  \caption{Binary datasets}
  \label{fig:gtt_boxplot_binary}
\end{subfigure}\hfill
\begin{subfigure}[t]{0.48\linewidth}
\setlength{\abovecaptionskip}{0pt}
 \centering
  \includegraphics[width=0.92\linewidth]{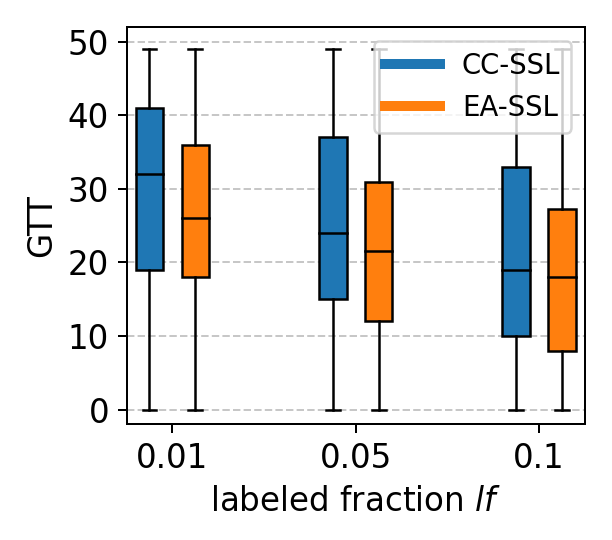}
  \caption{Multiclass datasets}
  \label{fig:gtt_boxplot_multiclass}
\end{subfigure}
\caption{Result distributions of GTT.}
\label{fig:gtt_dist}
\end{figure}

Figures~\ref{fig:gtt_boxplot_binary} and~\ref{fig:gtt_boxplot_multiclass} show the distributions of GTT. In contrast to TTT, GTT indicates faster convergence for EA-SSL in several settings, but the evidence depends on the significance threshold. On binary datasets, EA-SSL requires fewer generations than CC-SSL at $lf\in\{0.05,0.10\}$; the paired tests are significant at 95\% confidence ($p=0.0136$ and $p=0.0293$) but not at 99\% confidence ($p>0.01$). On multiclass datasets, EA-SSL reaches the target fitness in fewer generations at all labeled fractions; the paired tests are significant at 95\% confidence ($p<0.05$) but not at 99\% confidence ($p>0.01$).
These results indicate faster algorithmic convergence for EA-SSL in terms of evolutionary iterations.

%\paragraph{RQ4 (Cost): Is either approach preferable under a compute--performance trade-off measured with a runtime proxy?}Under the wall-clock proxy (TTT to target fitness), the results do not show a statistically significant difference between EA-SSL and CC-SSL ($p > 0.13$), so neither method is preferable by this proxy. Under the generational proxy (GTT), EA-SSL reaches the same target fitness in fewer generations overall ($p < 0.001$; median difference: two generations). These findings indicate similar time-to-quality in seconds, with fewer evolutionary steps for EA-SSL under the target-based criterion.

\textit{\textbf{RQ4:} Is either approach preferable under a compute--performance trade-off measured with a runtime proxy?}
Under the wall-clock proxy (TTT), CC-SSL and EA-SSL do not show systematic or practically meaningful differences, as reflected by overlapping boxplots and non-significant paired tests in most conditions.
Under the generational proxy (GTT), EA-SSL consistently reaches the target fitness in fewer generations, particularly on multiclass datasets.
These findings indicate that EA-SSL converges in fewer evolutionary steps, while both methods exhibit comparable time-to-quality in seconds.
Therefore, neither method is strictly preferable under a wall-clock cost criterion, while EA-SSL shows higher algorithmic convergence efficiency under a generation-based criterion.

%In general, the methods exhibit different trajectories under the same representation budget. EA-SSL shows higher median diversity and higher median best-so-far fitness and reaches the target fitness in fewer generations, while CC-SSL reduces diversity earlier. These differences do not correspond to systematic differences in final test performance, and both methods reach comparable test performance on these datasets.

In general, the methods exhibit different trajectories under the same representation budget. EA-SSL shows higher median diversity and best-so-far fitness and reaches the target fitness in fewer generations, while CC-SSL reduces diversity earlier. These differences do not correspond to systematic differences in final test performance, and both methods reach comparable test performance. % on these datasets.

\subsection{Pseudo-label dynamics and generalization diagnostics}\label{subsec:rq3_dynamics}
\begin{figure}[!h]
\setlength{\abovecaptionskip}{2pt}
\setlength{\belowcaptionskip}{0pt}
\centering
\begin{subfigure}[t]{0.48\linewidth}
\setlength{\abovecaptionskip}{0pt}

  \centering
  \includegraphics[width=0.92\linewidth]{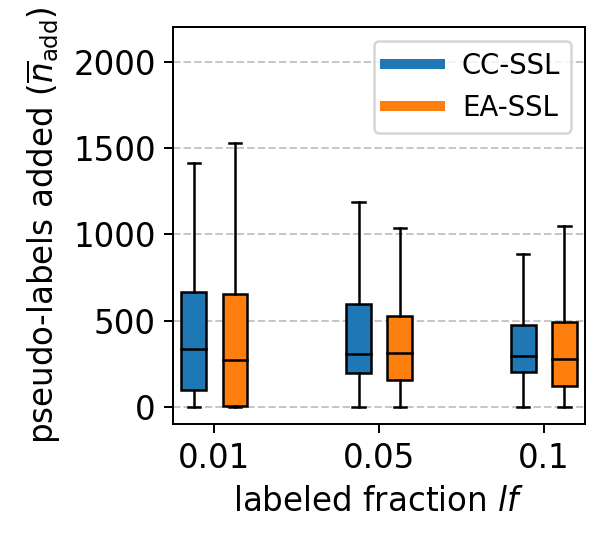}
  \caption{Binary datasets}
  \label{fig:pseudoaded_labels_binary}
\end{subfigure}\hfill
\begin{subfigure}[t]{0.48\linewidth}
\setlength{\abovecaptionskip}{0pt}

  \centering
  \includegraphics[width=0.92\linewidth]{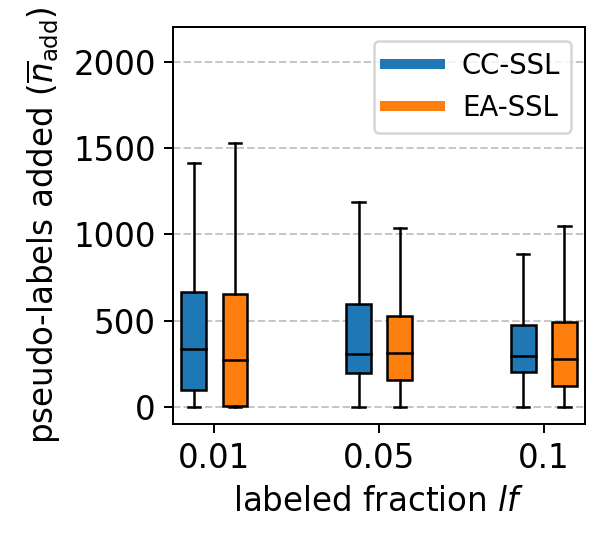}
  \caption{Multiclass datasets}
  \label{fig:pseudoaded_labels_multiclass}
\end{subfigure}
\caption{Distributions of pseudo-labels added}
\label{fig:pseudoaded_labels}
\end{figure}

\begin{figure}[!h]
\vspace{-0.2cm}
\setlength{\abovecaptionskip}{2pt}
\setlength{\belowcaptionskip}{-8pt}
\centering
\begin{subfigure}[t]{0.48\linewidth}
\setlength{\abovecaptionskip}{-1pt}
\setlength{\belowcaptionskip}{1pt}

  \centering
  \includegraphics[width=0.92\linewidth]{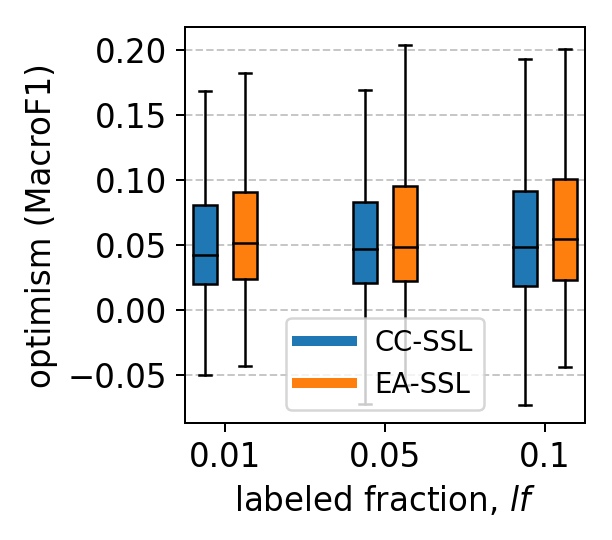}
  \caption{Binary datasets}
  \label{fig:optimism_binary}
\end{subfigure}\hfill
\begin{subfigure}[t]{0.48\linewidth}
\setlength{\abovecaptionskip}{-1pt}
\setlength{\belowcaptionskip}{1pt}

  \centering
  \includegraphics[width=0.92\linewidth]{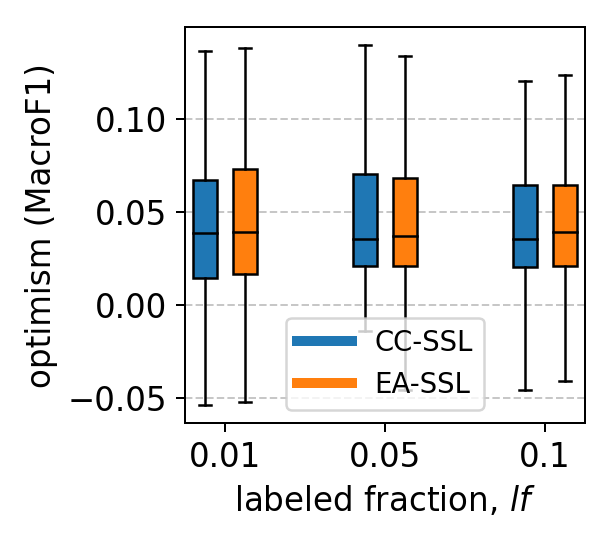}
  \caption{Multiclass datasets}
  \label{fig:optimism_multiclass}
\end{subfigure}
\caption{Distributions of optimism metric}
\label{fig:optimism}
\end{figure}

\vspace{-0.1cm}
This subsection tests whether CC-SSL and EA-SSL differ in (i) pseudo-label acceptance behavior, (ii) post-SSL stability on a held-out probe set, and (iii) validation--test generalization gaps.
Aggressiveness is measured by pseudo-label volume $\overline{n}_{\mathrm{add}}$ (Eq.~\eqref{eq:pseudo_added}), stability is measured by $\mathrm{ProbeDrop}$, and generalization is measured by MacroF1 validation optimism $\texttt{val\_minus\_test}$ defined in Section~\ref{sec:metrics}.
Figure~\ref{fig:pseudoaded_labels} summarizes pseudo-label volume and Figure~\ref{fig:optimism} summarizes optimism distributions for binary and multiclass datasets.

%This subsection addresses RQ3 by testing whether performance differences between CC-SSL and EA-SSL coincide with differences in pseudo-label dynamics or stability. Two metrics are analyzed: \emph{aggressiveness}, measured as the total number of pseudo-labels added during self-training (\texttt{pseudo\_added}), and a \emph{stability proxy}, measured as the change in performance on a held-out probe set after self-training relative to the initial model ($\mathrm{ProbeDrop}$). Positive $\mathrm{ProbeDrop}$ indicates probe degradation after training on pseudo-labels and is used as a proxy for confirmation bias (error amplification from incorrect pseudo-labels)~\cite{arazo2020pseudo,sohn2020fixmatch}.

Paired comparisons between CC-SSL and EA-SSL do not show statistically significant differences in the median values of pseudo-labels added ($\overline{n}_{\mathrm{add}}$) and $\mathrm{ProbeDrop}$ (Wilcoxon tests $p > 0.01$). The median paired difference in \texttt{pseudo\_added} remains close to zero across datasets, so the performance differences do not correspond to a systematic increase in pseudo-label volume.

Figure~\ref{fig:pseudoaded_labels} provides additional structure beyond the paired medians. On binary datasets (Figure~\ref{fig:pseudoaded_labels_binary}), CC-SSL and EA-SSL exhibit similar pseudo-label counts across labeled fractions. The main difference is dispersion: CC-SSL shows tighter distributions, while EA-SSL shows higher run-to-run variability, most prominently at $lf=0.01$. On multiclass datasets (Figure~\ref{fig:pseudoaded_labels_multiclass}), both methods add more pseudo-labels overall. EA-SSL shows higher pseudo-label counts at $lf \in \{0.05, 0.10\}$, while CC-SSL remains comparatively conservative. At $lf=0.01$, both methods show extreme variability, which is consistent with high sensitivity to early pseudo-label decisions in the multiclass low-label regime.

For stability, the median paired difference in $\mathrm{ProbeDrop}$ is $0.0$ (paired Wilcoxon; $p>0.01$). The $\mathrm{ProbeDrop}$ is zero in the large majority of runs: in the binary setting, $\mathrm{ProbeDrop}=0$ occurs in $96.5\%$ (CC-SSL) and $96.9\%$ (EA-SSL); and in the multiclass setting, $\mathrm{ProbeDrop}=0$ occurs in $96.85\%$ (CC-SSL) and $97.1\%$ (EA-SSL) of runs. This pattern is consistent with limited observable probe degradation under the applied pseudo-label selection policy~\cite{arazo2020pseudo,sohn2020fixmatch}. The proxy is conservative in this setting: a zero $\mathrm{ProbeDrop}$ can arise when model updates are small, pseudo-label noise is low, or the probe evaluation is insensitive to small changes.

Figure~\ref{fig:optimism} shows the distributions of MacroF1 validation optimism defined in Section~\ref{sec:metrics}.
Across both dataset types, CC-SSL and EA-SSL exhibit small positive median optimism at all labeled fractions, with overlapping interquartile ranges and occasional negative values.
Paired Wilcoxon tests do not detect statistically significant differences between CC-SSL and EA-SSL at any labeled fraction ($p>0.01$).
The observed gaps indicate mild optimistic validation estimates for both methods, with comparable magnitude and dispersion.
These results indicate similar validation--test gaps for CC-SSL and EA-SSL under the shared evaluation protocol, with limited evidence of method-dependent shifts in optimism.

%Overall, the diagnostics provide limited evidence that CC-SSL and EA-SSL differ systematically in pseudo-label dynamics or generalization behavior.
%Pseudo-label volume distributions overlap across labeled fractions (Figure~\ref{fig:pseudoaded_labels}), $\mathrm{ProbeDrop}$ is identically zero in more than $96\%$ of runs in both dataset types, and validation optimism shows overlapping distributions without significant paired differences (Figure~\ref{fig:optimism}; $p>0.01$).
%These outcomes indicate that the final-performance differences observed in Section~\ref{subsec:final_results} are not explained by consistent shifts in acceptance aggressiveness, probe-set degradation, or validation--test gaps under the shared pseudo-labeling protocol.

\textit{\textbf{RQ3:} When CC-SSL and EA-SSL differ in performance, do these differences co-occur with changes in pseudo-label dynamics, stability proxies, or validation--test optimism?}
No. Paired Wilcoxon tests do not detect differences between CC-SSL and EA-SSL in pseudo-label volume $\overline{n}_{\mathrm{add}}$, $\mathrm{ProbeDrop}$, or MacroF1 validation optimism at any labeled fraction ($p>0.01$). Pseudo-label volume distributions overlap (Figure~\ref{fig:pseudoaded_labels}), $\mathrm{ProbeDrop}=0$ in more than $96\%$ of runs, and optimism distributions overlap (Figure~\ref{fig:optimism}). These diagnostics do not support systematic method-dependent shifts in pseudo-label behavior or validation--test gaps under the shared protocol.

\section{Conclusions and Future Work}
\label{sec:conclusions}
This paper introduced CC-SSL, a lightweight cooperative coevolutionary framework that jointly adapts a two-view feature construction component and a pseudo-labeling policy for tabular semi-supervised classification, and compared it against a matched monolithic baseline (EA-SSL) and three lightweight SSL baselines under a staged tuning and paired-seed evaluation protocol.

Across 25 OpenML datasets and labeled fractions $\{1\%, 5\%, 10\%\}$, CC-SSL and EA-SSL achieved higher median test MacroF1 and accuracy than baselines, with the largest separations at $1\%$ labeled data. The benchmark did not show a systematic final-test advantage of cooperative factorization over monolithic encoding: most CC-SSL vs.\ EA-SSL comparisons were statistical draws on MacroF1 and accuracy under the paired tests reported in the analysis.

The methods differed in search behavior and convergence proxies. EA-SSL maintained higher median best-so-far fitness and higher population diversity across generations, while CC-SSL reduced diversity earlier; these differences did not translate into final-test gains. TTT results were comparable, whereas GTT favored EA-SSL in several conditions, particularly on multiclass datasets. Pseudo-label volume, $\mathrm{ProbeDrop}$, and validation optimism did not show statistically significant differences between CC-SSL and EA-SSL.
%, and $\mathrm{ProbeDrop}$ was zero in more than 96\% of runs in both binary and multiclass settings.

Future work will analyze EA-related design choices that can explain the observed fitness--diversity dynamics under the same evaluation budget, including alternative credit assignment between populations (decoupled attribution and marginal contribution), hybrid island--cooperative variants with controlled migration between CC instances~\cite{toutouh2023semi,hemberg2021spatial}, and diversity-aware survival.

\begin{acks}
This research is partially funded by the Universidad de M\'alaga (UMA) under grant number PID2024-158752OB-I00 (AIM-ZERO) by MICIU/AEI/10.13039/501100011033; Funding for open access charge: Universidad de M\'alaga / CBUA. The authors thank the Supercomputing and Bioinformatics center at the UMA for their computer resources and assistance.
\end{acks}

\bibliographystyle{plain}
\bibliography{ref}

\clearpage
\FloatBarrier % optional but helpful
\appendix

\section{Benchmark datasets}
\label{sec:datasets}

Table~\ref{tab:datasets} summarizes the benchmark datasets used in the experimental evaluation. All datasets are drawn from the OpenML-CC18 collection and cover binary and multiclass classification problems. The table reports the OpenML identifier, dataset name, number of observations ($n$), number of features ($p$), and number of classes ($cl$). The benchmark includes datasets with varying dimensionality and class cardinality, which allows evaluating performance and search behavior across different levels of class imbalance, feature dimensionality, and multiclass complexity under a unified protocol.

\begin{table}[!h]
\small
\centering
\caption{Benchmark datasets (OpenML-CC18) and basic characteristics. Rows = number of observations $n$, Feat. = number of features $p$, Classes = number of classes $cl$.}
\label{tab:datasets}
\setlength{\tabcolsep}{3pt}
\begin{tabular}{r l r r r}
\toprule
\textbf{ID} & \textbf{Name} & \textbf{Rows} & \textbf{Feat.} & \textbf{Classes}\\
\midrule
% -------- cl = 2 --------
3    & \texttt{kr-vs-kp}                    & 3196  & 37  & 2 \\
15   & \texttt{breast-w}                    & 699   & 10  & 2 \\
29   & \texttt{credit-approval}             & 690   & 16  & 2 \\
31   & \texttt{credit-g}                    & 1000  & 21  & 2 \\
37   & \texttt{diabetes}                    & 768   & 9   & 2 \\
38   & \texttt{sick}                        & 3772  & 30  & 2 \\
44   & \texttt{spambase}                    & 4601  & 58  & 2 \\
50   & \texttt{tic-tac-toe}                 & 958   & 10  & 2 \\
1049 & \texttt{pc4}                         & 1458  & 38  & 2 \\
1067 & \texttt{kc1}                         & 2109  & 22  & 2 \\
1480 & \texttt{ilpd}                        & 583   & 11  & 2 \\
\midrule
% -------- cl = 3 --------
11   & \texttt{balance-scale}               & 625   & 5   & 3 \\
23   & \texttt{cmc}                         & 1473  & 10  & 3 \\
46   & \texttt{splice}                      & 3190  & 62  & 3 \\
% -------- cl = 4 --------
54   & \texttt{vehicle}                     & 846   & 19  & 4 \\
% -------- cl = 9 --------
1468 & \texttt{cnae-9}                      & 1080  & 857 & 9 \\
% -------- cl = 10 --------
12   & \texttt{mfeat-factors}               & 2000  & 217 & 10 \\
14   & \texttt{mfeat-fourier}               & 2000  & 77  & 10 \\
16   & \texttt{mfeat-karhunen}              & 2000  & 65  & 10 \\
22   & \texttt{mfeat-zernike}               & 2000  & 48  & 10 \\
28   & \texttt{optdigits}                   & 5620  & 65  & 10 \\
% -------- cl = 11 --------
307  & \texttt{vowel}                       & 990   & 13  & 11 \\
\bottomrule
\end{tabular}
\end{table}

\section{Parameter Settings}
Table~\ref{tab:param_settings_staged} presents the configuration parameters explored in the parameter-setting experiments. In the first stage, the population sizes and the evolutionary budget are configured through tuples $(|popA|,|popB|,G)$, where $|popA|$ and $|popB|$ denote the population sizes of populations $A$ and $B$, respectively, and $G$ denotes the number of generations. In the second stage, the evolutionary operator probabilities are varied independently for each population: $p^{(A)}_{cx}$ and $p^{(A)}_{mut}$ correspond to the crossover and mutation probabilities of population $A$, while $p^{(B)}_{cx}$ and $p^{(B)}_{mut}$ correspond to the crossover and mutation probabilities of population $B$. In the third stage, the scalar-fitness aggregation is controlled by the weights $\lambda_{std}$, $\lambda_{bias}$, and $\lambda_{added}$, which penalize, respectively, the standard deviation of validation performance, the bias-related term, and the number of pseudo-labels added during training.

\begin{table}[!h]
\centering
\caption{Parameter configurations explored in the sensitivity analysis}
\label{tab:param_settings_staged}
\begin{tabular}{l r}
\hline
{Configuration parameter(s)} & {Tentative values} \\
\hline

\multicolumn{2}{c}{{Stage 1: Population sizes and generations}} \\
$(|popA|,|popB|,G)$
& $\{(6,6,50),\ (8,8,38),\ (10,10,30)\}$ \\
\hline

\multicolumn{2}{c}{{Stage 2: Evolutionary operator probabilities}} \\
$p^{(A)}_{cx}$
& $\{0.70,\ 0.85,\ 0.95\}$ \\
$p^{(A)}_{mut}$ 
& $\{0.25,\ 0.45,\ 0.65\}$ \\
$p^{(B)}_{cx}$
& $\{0.70,\ 0.85,\ 0.95\}$ \\
$p^{(B)}_{mut}$
& $\{0.35,\ 0.55,\ 0.75\}$ \\
\hline

\multicolumn{2}{c}{{Stage 3: Scalar-fitness weights}} \\
$\lambda_{std}$
& $\{0.0,\ 0.2,\ 0.4\}$ \\
$\lambda_{bias}$
& $\{0.3,\ 0.7,\ 1.1\}$ \\
$\lambda_{added}$
& $\{0.0,\ 0.0002,\ 0.0005\}$ \\
\hline
\end{tabular}
\end{table}

\section{Complementary results}
\label{sec:acc-results}

Tables~\ref{tab:final:acc-binary} and~\ref{tab:final:acc-multiclass} present descriptive statistics for test accuracy on the binary and multiclass benchmark datasets, respectively. Results are aggregated across datasets for each labeled fraction ($lf \in \{0.01, 0.05, 0.10\}$) and summarized using the minimum, median, interquartile range (IQR), and maximum accuracy. The column \emph{Wins} reports the number of datasets on which a method was statistically better according to Wilcoxon with Bonferroni correction.

\begin{table}[h]
\renewcommand{\arraystretch}{0.9}
\centering
\small
\caption{Descriptive statistics for accuracy results for 11~binary datasets} \label{tab:final:acc-binary}
\begin{tabular}{llrrrrr}
\toprule
$lf$ & Method & Min & Median & IQR & Max & Wins \\
\midrule
0.01 & CC-SSL & 0.632 & 0.836 & 0.179 & 0.963 & 5 \\
0.01 & EA-SSL & 0.639 & 0.836 & 0.189 & 0.966 & 5 \\
0.01 & \texttt{ST} & 0.537 & 0.667 & 0.234 & 0.943 & - \\
0.01 & \texttt{LS} & 0.522 & 0.654 & 0.233 & 0.939 & - \\
0.01 & \texttt{HCo} & 0.522 & 0.690 & 0.212 & 0.939 & - \\
0.01 & Best Regular & 0.653 & 0.714 & 0.172 & 0.952 & - \\
0.01 & Global Regular & 0.582 & 0.700 & 0.159 & 0.939 & - \\
\midrule
0.05 & CC-SSL & 0.674 & 0.855 & 0.198 & 0.966 & 3 \\
0.05 & EA-SSL & 0.667 & 0.865 & 0.202 & 0.966 & 4 \\
0.05 & \texttt{ST} & 0.653 & 0.783 & 0.153 & 0.945 & - \\
0.05 & \texttt{LS} & 0.444 & 0.700 & 0.186 & 0.939 & - \\
0.05 & \texttt{HCo} & 0.653 & 0.757 & 0.184 & 0.957 & - \\
0.05 & Best Regular & 0.674 & 0.836 & 0.190 & 0.962 & - \\
0.05 & Global Regular & 0.653 & 0.820 & 0.193 & 0.950 & - \\
\midrule
0.10 & CC-SSL & 0.686 & 0.846 & 0.195 & 0.948 & 3 \\
0.10 & EA-SSL & 0.686 & 0.845 & 0.192 & 0.952 & 2 \\
0.10 & \texttt{ST} & 0.669 & 0.838 & 0.174 & 0.943 & - \\
0.10 & \texttt{LS} & 0.522 & 0.676 & 0.115 & 0.879 & - \\
0.10 & \texttt{HCo} & 0.663 & 0.836 & 0.169 & 0.948 & - \\
0.10 & Best Regular & 0.703 & 0.850 & 0.153 & 0.964 & - \\
0.10 & Global Regular & 0.673 & 0.833 & 0.204 & 0.955 & - \\
\bottomrule
\end{tabular}
\end{table}

Tables~\ref{tab:full:f1-binray}--\ref{tab:full:acc-multiclass} provide dataset-wise results for MacroF1 and accuracy. Each cell reports the median (IQR) over 30~independent runs for a given dataset and labeled fraction. The tables also include two supervised baselines: Best Regular (BR), defined as the best-performing supervised classifier for that configuration, and Global Regular (GR), defined as the aggregated performance across all evaluated supervised classifiers. Shaded cells indicate cases where CC-SSL or EA-SSL significantly outperform all SSL baselines under a Bonferroni-corrected Wilcoxon test ($p<0.01$).

\begin{table}[tbp]
\renewcommand{\arraystretch}{0.9}
\centering
\small
\caption{Descriptive statistics for accuracy results for 11~multiclass datasets} \label{tab:final:acc-multiclass}
\begin{tabular}{llrrrrr}
\toprule
$lf$ & Method & Min & Median & IQR & Max & Wins \\
\midrule
0.01 & CC-SSL & 0.374 & 0.676 & 0.241 & 0.873 & 11 \\
0.01 & EA-SSL & 0.387 & 0.670 & 0.256 & 0.913 & 11 \\
0.01 & \texttt{ST} & 0.323 & 0.452 & 0.258 & 0.750 & - \\
0.01 & \texttt{LS} & 0.102 & 0.257 & 0.232 & 0.519 & - \\
0.01 & \texttt{HCo} & 0.316 & 0.473 & 0.173 & 0.678 & - \\
0.01 & Best Regular & 0.327 & 0.548 & 0.194 & 0.782 & - \\
0.01 & Global Regular & 0.293 & 0.487 & 0.173 & 0.728 & - \\
\midrule
0.05 & CC-SSL & 0.357 & 0.728 & 0.282 & 0.926 & 6 \\
0.05 & EA-SSL & 0.374 & 0.744 & 0.267 & 0.933 & 9 \\
0.05 & \texttt{ST} & 0.323 & 0.677 & 0.335 & 0.903 & - \\
0.05 & \texttt{LS} & 0.091 & 0.127 & 0.244 & 0.519 & - \\
0.05 & \texttt{HCo} & 0.296 & 0.628 & 0.237 & 0.896 & - \\
0.05 & Best Regular & 0.367 & 0.735 & 0.247 & 0.911 & - \\
0.05 & Global Regular & 0.301 & 0.651 & 0.233 & 0.880 & - \\
\midrule
0.10 & CC-SSL & 0.414 & 0.806 & 0.229 & 0.942 & 6 \\
0.10 & EA-SSL & 0.441 & 0.818 & 0.220 & 0.943 & 6 \\
0.10 & \texttt{ST} & 0.401 & 0.809 & 0.253 & 0.940 & - \\
0.10 & \texttt{LS} & 0.091 & 0.160 & 0.250 & 0.519 & - \\
0.10 & \texttt{HCo} & 0.354 & 0.756 & 0.230 & 0.923 & - \\
0.10 & Best Regular & 0.468 & 0.818 & 0.189 & 0.931 & - \\
0.10 & Global Regular & 0.384 & 0.747 & 0.211 & 0.923 & - \\
\bottomrule
\end{tabular}
\end{table}

\begin{table*}[!h]
\setlength{\tabcolsep}{2pt}
\renewcommand{\arraystretch}{0.9}
\small
\centering
\caption{MacroF1 results comparison for 11 binary datasets, median(IQR). Shaded: CC-SSL/EA-SSL significantly better than all baselines, Bonferroni-corrected Wilcoxon, p < 0.01}\label{tab:full:f1-binray}
\small
\begin{tabular}{lccccccccccc}
\toprule
Method & 3 & 15 & 29 & 31 & 37 & 38 & 44 & 50 & 1049 & 1067 & 1480 \\
\midrule
\multicolumn{12}{c}{$lf$=0.01} \\
\midrule
CC-SSL & \cellcolor{gray!20}0.850(0.026) & 0.948(0.020) & \cellcolor{gray!20}0.835(0.069) & \cellcolor{gray!20}0.540(0.063) & \cellcolor{gray!20}0.675(0.049) & \cellcolor{gray!20}0.842(0.041) & \cellcolor{gray!20}0.893(0.016) & 0.561(0.073) & 0.650(0.052) & 0.652(0.034) & \cellcolor{gray!20}0.612(0.044) \\
EA-SSL & \cellcolor{gray!20}0.876(0.030) & 0.953(0.010) & \cellcolor{gray!20}0.835(0.042) & \cellcolor{gray!20}0.579(0.046) & \cellcolor{gray!20}0.694(0.069) & \cellcolor{gray!20}0.851(0.026) & \cellcolor{gray!20}0.897(0.028) & \cellcolor{gray!20}0.579(0.068) & 0.657(0.078) & 0.649(0.032) & 0.608(0.046) \\
\texttt{ST} & 0.450(0.292) & 0.936(0.066) & 0.357(0.070) & 0.412(0.091) & 0.445(0.191) & 0.484(0.048) & 0.668(0.243) & 0.530(0.068) & 0.587(0.089) & 0.608(0.148) & 0.417(0.135) \\
\texttt{LS} & 0.343(0.020) & 0.760(0.145) & 0.357(0.070) & 0.412(0.000) & 0.426(0.155) & 0.484(0.000) & 0.377(0.002) & 0.395(0.000) & 0.468(0.014) & 0.503(0.083) & 0.421(0.214) \\
\texttt{HCo} & 0.343(0.314) & 0.912(0.098) & 0.375(0.250) & 0.412(0.101) & 0.407(0.081) & 0.496(0.048) & 0.592(0.345) & 0.516(0.075) & 0.597(0.117) & 0.639(0.064) & 0.484(0.165) \\
BR & 0.713(0.058) & 0.948(0.032) & 0.675(0.177) & 0.522(0.066) & 0.610(0.093) & 0.749(0.073) & 0.825(0.064) & 0.553(0.034) & 0.601(0.056) & 0.632(0.099) & 0.568(0.122) \\
GR & 0.692(0.235) & 0.916(0.356) & 0.562(0.316) & 0.497(0.099) & 0.580(0.232) & 0.665(0.205) & 0.763(0.323) & 0.527(0.153) & 0.569(0.135) & 0.578(0.184) & 0.514(0.174) \\
\midrule
\multicolumn{12}{c}{$lf$=0.01} \\
\midrule
CC-SSL & \cellcolor{gray!20}0.920(0.015) & 0.943(0.027) & \cellcolor{gray!20}0.854(0.030) & 0.600(0.090) & 0.696(0.043) & \cellcolor{gray!20}0.858(0.041) & \cellcolor{gray!20}0.898(0.009) & \cellcolor{gray!20}0.639(0.073) & \cellcolor{gray!20}0.677(0.044) & \cellcolor{gray!20}0.647(0.021) & \cellcolor{gray!20}0.593(0.079) \\
EA-SSL & \cellcolor{gray!20}0.931(0.015) & 0.953(0.022) & \cellcolor{gray!20}0.864(0.036) & 0.606(0.078) & \cellcolor{gray!20}0.711(0.041) & 0.866(0.025) & \cellcolor{gray!20}0.902(0.012) & \cellcolor{gray!20}0.595(0.095) & \cellcolor{gray!20}0.668(0.065) & \cellcolor{gray!20}0.650(0.015) & \cellcolor{gray!20}0.571(0.067) \\
\texttt{ST} & 0.832(0.064) & 0.936(0.027) & 0.768(0.258) & 0.565(0.083) & 0.657(0.058) & 0.600(0.149) & 0.743(0.125) & 0.437(0.203) & 0.538(0.075) & 0.530(0.082) & 0.460(0.081) \\
\texttt{LS} & 0.343(0.020) & 0.694(0.207) & 0.308(0.049) & 0.412(0.000) & 0.394(0.000) & 0.484(0.000) & 0.379(0.003) & 0.395(0.000) & 0.468(0.014) & 0.496(0.038) & 0.417(0.000) \\
\texttt{HCo} & 0.739(0.241) & 0.953(0.011) & 0.650(0.210) & 0.528(0.178) & 0.478(0.187) & 0.484(0.000) & 0.851(0.056) & 0.517(0.161) & 0.551(0.089) & 0.543(0.145) & 0.453(0.155) \\
BR & 0.911(0.040) & 0.957(0.011) & 0.829(0.067) & 0.598(0.051) & 0.692(0.044) & 0.789(0.077) & 0.901(0.006) & 0.608(0.032) & 0.649(0.079) & 0.615(0.046) & 0.585(0.096) \\
GR & 0.885(0.094) & 0.945(0.217) & 0.769(0.230) & 0.564(0.158) & 0.665(0.154) & 0.710(0.166) & 0.874(0.038) & 0.570(0.161) & 0.614(0.091) & 0.597(0.088) & 0.548(0.164) \\
\midrule
\multicolumn{12}{c}{$lf$=0.01} \\
\midrule
CC-SSL & \cellcolor{gray!20}0.939(0.012) & 0.942(0.022) & 0.850(0.022) & \cellcolor{gray!20}0.605(0.042) & \cellcolor{gray!20}0.682(0.056) & 0.799(0.058) & \cellcolor{gray!20}0.909(0.009) & 0.660(0.039) & \cellcolor{gray!20}0.694(0.084) & \cellcolor{gray!20}0.658(0.054) & 0.606(0.092) \\
EA-SSL & \cellcolor{gray!20}0.944(0.009) & 0.948(0.020) & 0.855(0.026) & 0.610(0.054) & 0.691(0.048) & 0.780(0.062) & \cellcolor{gray!20}0.908(0.011) & 0.658(0.043) & \cellcolor{gray!20}0.702(0.039) & \cellcolor{gray!20}0.663(0.017) & 0.577(0.089) \\
\texttt{ST} & 0.896(0.051) & 0.935(0.029) & 0.818(0.039) & 0.495(0.192) & 0.666(0.159) & 0.789(0.137) & 0.835(0.068) & 0.609(0.114) & 0.568(0.098) & 0.567(0.149) & 0.553(0.088) \\
\texttt{LS} & 0.343(0.000) & 0.608(0.141) & 0.357(0.000) & 0.412(0.000) & 0.394(0.000) & 0.784(0.098) & 0.379(0.005) & 0.395(0.000) & 0.468(0.014) & 0.479(0.039) & 0.417(0.000) \\
\texttt{HCo} & 0.873(0.062) & 0.942(0.022) & 0.821(0.068) & 0.422(0.140) & 0.588(0.154) & 0.745(0.153)  & 0.838(0.039) & 0.518(0.156) & 0.531(0.127) & 0.543(0.120) & 0.463(0.135) \\
BR & 0.942(0.017) & 0.953(0.011) & 0.840(0.049) & 0.602(0.053) & 0.684(0.035) & 0.803(0.036) & 0.916(0.005) & 0.799(0.170) & 0.705(0.038) & 0.632(0.061) & 0.597(0.064) \\
GR & 0.938(0.023) & 0.937(0.178) & 0.803(0.198) & 0.578(0.052) & 0.674(0.047) & 0.783(0.086) & 0.901(0.036) & 0.664(0.085) & 0.676(0.082) & 0.609(0.069) & 0.553(0.152) \\
\bottomrule
\end{tabular}
\end{table*}

\begin{table*}[!h]
\setlength{\tabcolsep}{2pt}
\renewcommand{\arraystretch}{0.9}
\small
\centering
\caption{MacroF1 results comparison for 11 multiclass datasets, median(IQR). Shaded: CC-SSL/EA-SSL significantly better than all baselines, Bonferroni-corrected Wilcoxon, p < 0.01}\label{tab:full:f1-multiclass}
\small
\begin{tabular}{lccccccccccc}
\toprule
Method & 11 & 12 & 14 & 16 & 22 & 23 & 28 & 46 & 54 & 307 & 1468 \\
\midrule
\multicolumn{12}{c}{$lf$=0.01} \\
\midrule
\midrule
CC-SSL & \cellcolor{gray!20}0.538(0.031) & \cellcolor{gray!20}0.872(0.065) & \cellcolor{gray!20}0.563(0.054) & \cellcolor{gray!20}0.728(0.095) & \cellcolor{gray!20}0.615(0.048) & \cellcolor{gray!20}0.418(0.069) & \cellcolor{gray!20}0.823(0.090) & \cellcolor{gray!20}0.855(0.235) & \cellcolor{gray!20}0.503(0.060) & \cellcolor{gray!20}0.368(0.040) & \cellcolor{gray!20}0.688(0.062) \\
EA-SSL & \cellcolor{gray!20}0.539(0.039) & \cellcolor{gray!20}0.913(0.018) & \cellcolor{gray!20}0.593(0.070) & \cellcolor{gray!20}0.745(0.125) & \cellcolor{gray!20}0.628(0.037) & \cellcolor{gray!20}0.409(0.084) & \cellcolor{gray!20}0.881(0.036) & \cellcolor{gray!20}0.864(0.047) & \cellcolor{gray!20}0.521(0.075) & \cellcolor{gray!20}0.387(0.072) & \cellcolor{gray!20}0.687(0.066) \\
\texttt{ST} & 0.406(0.088) & 0.709(0.061) & 0.329(0.188) & 0.621(0.130) & 0.399(0.101) & 0.353(0.132) & 0.532(0.159) & 0.599(0.389) & 0.313(0.078) & 0.317(0.061) & 0.318(0.149) \\
\texttt{LS} & 0.383(0.307) & 0.346(0.166) & 0.118(0.093) & 0.077(0.036) & 0.186(0.056) & 0.171(0.065) & 0.022(0.005) & 0.228(0.099) & 0.259(0.090) & 0.047(0.061) & 0.022(0.000) \\
\texttt{HCo} & 0.406(0.072) & 0.661(0.100) & 0.463(0.077) & 0.619(0.093) & 0.407(0.105) & 0.353(0.095) & 0.519(0.141) & 0.565(0.358) & 0.386(0.193) & 0.295(0.063) & 0.487(0.161) \\
BR & 0.456(0.088) & 0.777(0.048) & 0.484(0.062) & 0.633(0.082) & 0.558(0.077) & 0.377(0.076) & 0.746(0.061) & 0.616(0.063) & 0.426(0.077) & 0.320(0.042) & 0.659(0.043) \\
GR & 0.399(0.215) & 0.726(0.269) & 0.439(0.206) & 0.541(0.278) & 0.475(0.234) & 0.337(0.135) & 0.667(0.323) & 0.538(0.329) & 0.388(0.224) & 0.292(0.161) & 0.564(0.265) \\
\midrule
\multicolumn{12}{c}{$lf$=0.05} \\
\midrule
CC-SSL & 0.599(0.040) & \cellcolor{gray!20}0.923(0.017) & \cellcolor{gray!20}0.623(0.068) & \cellcolor{gray!20}0.858(0.036) & \cellcolor{gray!20}0.716(0.050) & 0.421(0.063) & \cellcolor{gray!20}0.920(0.015) & 0.920(0.018) & \cellcolor{gray!20}0.559(0.072) & \cellcolor{gray!20}0.358(0.043) & \cellcolor{gray!20}0.731(0.057) \\
EA-SSL & \cellcolor{gray!20}0.609(0.058) & \cellcolor{gray!20}0.934(0.014) & \cellcolor{gray!20}0.652(0.083) & \cellcolor{gray!20}0.873(0.028) & \cellcolor{gray!20}0.696(0.018) & \cellcolor{gray!20}0.420(0.050) & \cellcolor{gray!20}0.924(0.010) & \cellcolor{gray!20}0.920(0.015) & \cellcolor{gray!20}0.592(0.050) & 0.354(0.043) & \cellcolor{gray!20}0.744(0.057) \\
\texttt{ST} & 0.584(0.031) & 0.893(0.025) & 0.539(0.088) & 0.788(0.103) & 0.642(0.062) & 0.375(0.071) & 0.904(0.021) & 0.843(0.020) & 0.413(0.069) & 0.320(0.049) & 0.572(0.153) \\
\texttt{LS} & 0.211(0.014) & 0.054(0.061) & 0.025(0.036) & 0.018(0.000) & 0.043(0.049) & 0.200(0.028) & 0.028(0.057) & 0.228(0.000) & 0.168(0.097) & 0.015(0.000) & 0.022(0.000) \\
\texttt{HCo} & 0.493(0.066) & 0.877(0.038) & 0.529(0.082) & 0.765(0.064) & 0.620(0.049) & 0.350(0.066) & 0.897(0.015) & 0.742(0.119) & 0.517(0.076) & 0.279(0.041) & 0.606(0.082) \\
BR & 0.628(0.060) & 0.877(0.033) & 0.593(0.044) & 0.779(0.032) & 0.669(0.028) & 0.408(0.067) & 0.912(0.009) & 0.879(0.027) & 0.596(0.039) & 0.352(0.064) & 0.731(0.034) \\
GR & 0.551(0.281) & 0.825(0.075) & 0.552(0.071) & 0.696(0.109) & 0.608(0.124) & 0.394(0.076) & 0.880(0.041) & 0.828(0.078) & 0.520(0.258) & 0.292(0.182) & 0.651(0.306) \\
\midrule
\multicolumn{12}{c}{$lf$=0.10} \\
\midrule
CC-SSL & 0.609(0.026) & \cellcolor{gray!20}0.942(0.017) & \cellcolor{gray!20}0.722(0.024) & \cellcolor{gray!20}0.892(0.025) & \cellcolor{gray!20}0.754(0.012) & \cellcolor{gray!20}0.437(0.054) & 0.936(0.010) & 0.915(0.024) & \cellcolor{gray!20}0.636(0.048) & 0.395(0.061) & 0.807(0.053) \\
EA-SSL & 0.630(0.040) & \cellcolor{gray!20}0.943(0.012) & \cellcolor{gray!20}0.733(0.036) & \cellcolor{gray!20}0.907(0.014) & \cellcolor{gray!20}0.747(0.028) & \cellcolor{gray!20}0.438(0.040) & 0.935(0.010) & 0.920(0.014) & \cellcolor{gray!20}0.649(0.042) & 0.413(0.069) & 0.818(0.058) \\
\texttt{ST} & 0.609(0.025) & 0.925(0.025) & 0.673(0.039) & 0.865(0.023) & 0.709(0.041) & 0.344(0.103) & 0.929(0.008) & 0.935(0.015) & 0.589(0.084) & 0.383(0.026) & 0.802(0.054) \\
\texttt{LS} & 0.211(0.218) & 0.078(0.140) & 0.018(0.058) & 0.018(0.074) & 0.088(0.108) & 0.200(0.000) & 0.029(0.066) & 0.228(0.000) & 0.173(0.128) & 0.015(0.018) & 0.022(0.000) \\
\texttt{HCo} & 0.563(0.069) & 0.916(0.027) & 0.653(0.044) & 0.845(0.036) & 0.721(0.051) & 0.303(0.150) & 0.923(0.010) & 0.842(0.054) & 0.580(0.061) & 0.328(0.033) & 0.758(0.086) \\
BR & 0.617(0.062) & 0.926(0.015) & 0.728(0.041) & 0.855(0.023) & 0.730(0.035) & 0.441(0.045) & 0.931(0.008) & 0.920(0.012) & 0.662(0.057) & 0.445(0.033) & 0.814(0.049) \\
GR & 0.595(0.185) & 0.904(0.029) & 0.664(0.077) & 0.822(0.061) & 0.699(0.079) & 0.435(0.043) & 0.923(0.025) & 0.878(0.060) & 0.615(0.101) & 0.373(0.060) & 0.749(0.286) \\
\bottomrule
\end{tabular}
\end{table*}

\begin{table*}[!h]
\setlength{\tabcolsep}{2pt}
\renewcommand{\arraystretch}{0.9}
\small
\centering
\caption{Accuracy results comparison for 11 binary datasets, median(IQR). Shaded: CC-SSL/EA-SSL significantly better than all baselines, Bonferroni-corrected Wilcoxon, p < 0.01}\label{tab:full:acc-binray}
\small
\begin{tabular}{lccccccccccc}
\toprule
Method & 3 & 15 & 29 & 31 & 37 & 38 & 44 & 50 & 1049 & 1067 & 1480 \\
\midrule
\multicolumn{12}{c}{$lf$=0.01} \\
\midrule
CC-SSL & \cellcolor{gray!20}0.852(0.025) & 0.952(0.019) & \cellcolor{gray!20}0.836(0.063) & 0.643(0.068) & \cellcolor{gray!20}0.719(0.063) & \cellcolor{gray!20}0.963(0.007) & \cellcolor{gray!20}0.898(0.016) & 0.632(0.071) & 0.854(0.039) & 0.820(0.035) & 0.674(0.029) \\
EA-SSL & \cellcolor{gray!20}0.877(0.029) & 0.957(0.010) & \cellcolor{gray!20}0.836(0.041) & 0.667(0.041) & \cellcolor{gray!20}0.732(0.056) & \cellcolor{gray!20}0.966(0.007) & \cellcolor{gray!20}0.902(0.026) & 0.639(0.057) & 0.877(0.041) & 0.825(0.021) & 0.669(0.051) \\
\texttt{ST} & 0.537(0.161) & 0.943(0.051) & 0.556(0.120) & 0.667(0.343) & 0.602(0.223) & 0.937(0.005) & 0.677(0.196) & 0.597(0.055) & 0.829(0.070) & 0.837(0.037) & 0.663(0.106) \\
\texttt{LS} & 0.522(0.045) & 0.819(0.087) & 0.556(0.120) & 0.700(0.000) & 0.654(0.043) & 0.939(0.000) & 0.606(0.001) & 0.653(0.000) & 0.879(0.000) & 0.847(0.003) & 0.594(0.366) \\
\texttt{HCo} & 0.522(0.165) & 0.924(0.076) & 0.522(0.193) & 0.690(0.150) & 0.649(0.163) & 0.939(0.004) & 0.695(0.202) & 0.615(0.053) & 0.815(0.086) & 0.836(0.036) & 0.611(0.160) \\
BR & 0.714(0.057) & 0.952(0.026) & 0.681(0.162) & 0.700(0.000) & 0.654(0.069) & 0.942(0.006) & 0.846(0.059) & 0.653(0.000) & 0.879(0.000) & 0.845(0.000) & 0.714(0.000) \\
GR & 0.700(0.192) & 0.926(0.215) & 0.582(0.239) & 0.668(0.128) & 0.649(0.032) & 0.939(0.008) & 0.787(0.224) & 0.608(0.085) & 0.777(0.170) & 0.814(0.060) & 0.634(0.146) \\
\midrule
\multicolumn{12}{c}{{$lf$=0.05}} \\
\midrule
CC-SSL & \cellcolor{gray!20}0.920(0.016) & 0.948(0.024) & \cellcolor{gray!20}0.855(0.029) & 0.693(0.038) & 0.732(0.050) & 0.966(0.010) & \cellcolor{gray!20}0.902(0.010) & 0.694(0.035) & 0.888(0.014) & 0.839(0.017) & 0.674(0.026) \\
EA-SSL & \cellcolor{gray!20}0.931(0.015) & 0.957(0.019) & \cellcolor{gray!20}0.865(0.034) & 0.687(0.018) & \cellcolor{gray!20}0.745(0.030) & 0.966(0.005) & \cellcolor{gray!20}0.905(0.013) & 0.667(0.045) & 0.884(0.025) & 0.825(0.046) & 0.669(0.043) \\
\texttt{ST} & 0.832(0.062) & 0.943(0.023) & 0.783(0.205) & 0.698(0.068) & 0.714(0.051) & 0.945(0.009) & 0.783(0.079) & 0.653(0.024) & 0.881(0.009) & 0.848(0.007) & 0.709(0.039) \\
\texttt{LS} & 0.522(0.045) & 0.781(0.111) & 0.444(0.111) & 0.700(0.000) & 0.649(0.000) & 0.939(0.000) & 0.607(0.001) & 0.653(0.000) & 0.879(0.002) & 0.847(0.003) & 0.714(0.000) \\
\texttt{HCo} & 0.757(0.201) & 0.957(0.010) & 0.691(0.173) & 0.700(0.029) & 0.662(0.057) & 0.939(0.000) & 0.866(0.048) & 0.653(0.018) & 0.879(0.010) & 0.845(0.009) & 0.686(0.071) \\
BR & 0.911(0.038) & 0.962(0.010) & 0.836(0.068) & 0.700(0.000) & 0.723(0.041) & 0.954(0.013) & 0.907(0.006) & 0.674(0.047) & 0.877(0.017) & 0.831(0.036) & 0.714(0.000) \\
GR & 0.886(0.090) & 0.950(0.137) & 0.775(0.185) & 0.687(0.043) & 0.695(0.075) & 0.949(0.018) & 0.881(0.037) & 0.653(0.029) & 0.862(0.037) & 0.820(0.037) & 0.669(0.091) \\
\midrule
\multicolumn{12}{c}{{$lf$=0.10}} \\
\midrule
CC-SSL & \cellcolor{gray!20}0.939(0.012) & 0.948(0.019) & 0.850(0.022) & 0.697(0.035) & \cellcolor{gray!20}0.745(0.019) & 0.962(0.010) & \cellcolor{gray!20}0.912(0.008) & 0.701(0.033) & 0.893(0.017) & 0.842(0.017) & 0.686(0.037) \\
EA-SSL & \cellcolor{gray!20}0.944(0.009) & 0.952(0.017) & 0.857(0.023) & 0.697(0.035) & 0.745(0.028) & 0.964(0.012) & \cellcolor{gray!20}0.912(0.010) & 0.705(0.040) & 0.888(0.021) & 0.833(0.028) & 0.686(0.043) \\
\texttt{ST} & 0.896(0.050) & 0.943(0.026) & 0.826(0.043) & 0.700(0.013) & 0.719(0.060) & 0.958(0.007) & 0.850(0.051) & 0.694(0.046) & 0.888(0.011) & 0.850(0.012) & 0.669(0.080) \\
\texttt{LS} & 0.522(0.000) & 0.738(0.064) & 0.556(0.000) & 0.700(0.000) & 0.649(0.000) & 0.962(0.009) & 0.607(0.001) & 0.653(0.000) & 0.879(0.002) & 0.847(0.003) & 0.714(0.000) \\
\texttt{HCo} & 0.875(0.060) & 0.948(0.019) & 0.821(0.068) & 0.700(0.000) & 0.697(0.068) & 0.958(0.012) & 0.854(0.030) & 0.663(0.025) & 0.884(0.011) & 0.850(0.007) & 0.703(0.030) \\
BR & 0.943(0.017) & 0.957(0.010) & 0.841(0.048) & 0.703(0.030) & 0.732(0.028) & 0.964(0.010) & 0.921(0.004) & 0.826(0.156) & 0.886(0.021) & 0.850(0.018) & 0.714(0.000) \\
GR & 0.938(0.023) & 0.943(0.107) & 0.804(0.144) & 0.673(0.051) & 0.719(0.037) & 0.955(0.015) & 0.905(0.035) & 0.715(0.063) & 0.873(0.025) & 0.833(0.036) & 0.686(0.069) \\
\bottomrule
\end{tabular}
\end{table*}

\begin{table*}[!h]
\setlength{\tabcolsep}{2pt}
\renewcommand{\arraystretch}{0.9}
\small
\centering
\caption{Accuracy results comparison for 11 multiclass datasets, median(IQR). Shaded: CC-SSL/EA-SSL significantly better than all baselines, Bonferroni-corrected Wilcoxon, p < 0.01}\label{tab:full:acc-multiclass}
\small
\begin{tabular}{lccccccccccc}
\toprule
Method & 11 & 12 & 14 & 16 & 22 & 23 & 28 & 46 & 54 & 307 & 1468 \\
\midrule
\multicolumn{12}{c}{$lf$=0.01} \\
\midrule
CC-SSL & \cellcolor{gray!20}0.676(0.043) & \cellcolor{gray!20}0.873(0.067) & \cellcolor{gray!20}0.567(0.053) & \cellcolor{gray!20}0.728(0.095) & \cellcolor{gray!20}0.620(0.049) & \cellcolor{gray!20}0.432(0.031) & \cellcolor{gray!20}0.825(0.080) & \cellcolor{gray!20}0.864(0.183) & \cellcolor{gray!20}0.504(0.045) & \cellcolor{gray!20}0.374(0.044) & \cellcolor{gray!20}0.682(0.048) \\
EA-SSL & \cellcolor{gray!20}0.670(0.066) & \cellcolor{gray!20}0.913(0.018) & \cellcolor{gray!20}0.593(0.062) & \cellcolor{gray!20}0.753(0.131) & \cellcolor{gray!20}0.630(0.030) & \cellcolor{gray!20}0.421(0.036) & \cellcolor{gray!20}0.881(0.037) & \cellcolor{gray!20}0.873(0.044) & \cellcolor{gray!20}0.520(0.071) & \cellcolor{gray!20}0.387(0.067) & \cellcolor{gray!20}0.688(0.074) \\
\texttt{ST} & 0.457(0.109) & 0.750(0.067) & 0.423(0.110) & 0.630(0.123) & 0.452(0.070) & 0.369(0.064) & 0.606(0.141) & 0.635(0.154) & 0.350(0.080) & 0.323(0.065) & 0.340(0.191) \\
\texttt{LS} & 0.489(0.420) & 0.408(0.190) & 0.197(0.101) & 0.177(0.064) & 0.257(0.082) & 0.346(0.111) & 0.102(0.003) & 0.519(0.279) & 0.343(0.054) & 0.114(0.055) & 0.111(0.000) \\
\texttt{HCo} & 0.452(0.142) & 0.678(0.086) & 0.473(0.062) & 0.618(0.093) & 0.432(0.082) & 0.373(0.074) & 0.557(0.115) & 0.630(0.133) & 0.398(0.121) & 0.316(0.072) & 0.512(0.181) \\
BR & 0.495(0.101) & 0.782(0.041) & 0.490(0.045) & 0.638(0.080) & 0.548(0.079) & 0.428(0.081) & 0.749(0.062) & 0.650(0.054) & 0.445(0.093) & 0.327(0.034) & 0.673(0.034) \\
GR & 0.463(0.097) & 0.728(0.251) & 0.453(0.190) & 0.545(0.256) & 0.487(0.208) & 0.378(0.092) & 0.679(0.291) & 0.620(0.114) & 0.394(0.182) & 0.293(0.141) & 0.573(0.249) \\
\midrule
\multicolumn{12}{c}{$lf$=0.05} \\
\midrule
CC-SSL & 0.851(0.066) & \cellcolor{gray!20}0.923(0.017) & \cellcolor{gray!20}0.643(0.071) & \cellcolor{gray!20}0.858(0.033) & \cellcolor{gray!20}0.718(0.048) & 0.459(0.046) & 0.921(0.015) & 0.926(0.018) & 0.571(0.071) & \cellcolor{gray!20}0.357(0.067) & \cellcolor{gray!20}0.728(0.049) \\
EA-SSL & 0.851(0.051) & \cellcolor{gray!20}0.933(0.014) & \cellcolor{gray!20}0.662(0.080) & \cellcolor{gray!20}0.873(0.027) & \cellcolor{gray!20}0.702(0.015) & 0.464(0.043) & \cellcolor{gray!20}0.924(0.009) & \cellcolor{gray!20}0.928(0.013) & \cellcolor{gray!20}0.602(0.075) & \cellcolor{gray!20}0.374(0.047) & \cellcolor{gray!20}0.744(0.056) \\
\texttt{ST} & 0.840(0.047) & 0.892(0.024) & 0.598(0.076) & 0.797(0.074) & 0.677(0.060) & 0.437(0.041) & 0.903(0.021) & 0.864(0.021) & 0.433(0.082) & 0.323(0.046) & 0.617(0.153) \\
\texttt{LS} & 0.463(0.009) & 0.127(0.095) & 0.103(0.049) & 0.100(0.000) & 0.132(0.048) & 0.428(0.081) & 0.104(0.061) & 0.519(0.000) & 0.268(0.075) & 0.091(0.000) & 0.111(0.000) \\
\texttt{HCo} & 0.713(0.093) & 0.877(0.034) & 0.558(0.069) & 0.772(0.051) & 0.628(0.040) & 0.434(0.036) & 0.896(0.015) & 0.784(0.077) & 0.524(0.115) & 0.296(0.049) & 0.611(0.120) \\
BR & 0.830(0.069) & 0.877(0.032) & 0.615(0.057) & 0.782(0.029) & 0.677(0.031) & 0.468(0.046) & 0.911(0.009) & 0.894(0.021) & 0.598(0.055) & 0.367(0.076) & 0.735(0.022) \\
GR & 0.726(0.363) & 0.828(0.075) & 0.565(0.067) & 0.702(0.104) & 0.621(0.103) & 0.437(0.059) & 0.880(0.042) & 0.844(0.065) & 0.522(0.215) & 0.301(0.173) & 0.651(0.302) \\
\midrule
\multicolumn{12}{c}{$lf$=0.10} \\
\midrule
CC-SSL & 0.851(0.045) & \cellcolor{gray!20}0.942(0.017) & \cellcolor{gray!20}0.722(0.029) & \cellcolor{gray!20}0.892(0.026) & \cellcolor{gray!20}0.753(0.016) & \cellcolor{gray!20}0.482(0.040) & 0.937(0.010) & 0.922(0.020) & \cellcolor{gray!20}0.634(0.051) & 0.414(0.077) & 0.806(0.056) \\
EA-SSL & 0.867(0.029) & \cellcolor{gray!20}0.943(0.012) & \cellcolor{gray!20}0.733(0.038) & \cellcolor{gray!20}0.907(0.014) & \cellcolor{gray!20}0.745(0.028) & 0.489(0.031) & 0.935(0.011) & 0.927(0.015) & \cellcolor{gray!20}0.661(0.028) & \cellcolor{gray!20}0.441(0.086) & 0.818(0.057) \\
\texttt{ST} & 0.851(0.032) & 0.925(0.024) & 0.682(0.020) & 0.865(0.023) & 0.728(0.035) & 0.459(0.031) & 0.929(0.008) & 0.940(0.011) & 0.602(0.062) & 0.401(0.027) & 0.809(0.044) \\
\texttt{LS} & 0.463(0.178) & 0.190(0.178) & 0.100(0.072) & 0.100(0.063) & 0.160(0.132) & 0.428(0.000) & 0.107(0.097) & 0.519(0.000) & 0.280(0.082) & 0.091(0.015) & 0.111(0.000) \\
\texttt{HCo} & 0.809(0.096) & 0.917(0.026) & 0.662(0.025) & 0.847(0.040) & 0.727(0.042) & 0.434(0.032) & 0.923(0.010) & 0.866(0.045) & 0.591(0.053) & 0.354(0.034) & 0.756(0.080) \\
BR & 0.856(0.035) & 0.927(0.014) & 0.737(0.034) & 0.855(0.022) & 0.732(0.036) & 0.473(0.044) & 0.931(0.008) & 0.928(0.011) & 0.673(0.041) & 0.468(0.045) & 0.818(0.049) \\
GR & 0.782(0.246) & 0.904(0.030) & 0.670(0.074) & 0.823(0.060) & 0.696(0.068) & 0.466(0.037) & 0.923(0.025) & 0.890(0.054) & 0.622(0.097) & 0.384(0.063) & 0.747(0.238) \\
\bottomrule
\end{tabular}
\end{table*}

\begin{table*}[!t]
\centering
\caption{Genotype specification shared by CC-SSL and EA-SSL. CC-SSL evolves $a$ and $b$ in separate populations; EA-SSL concatenates the same genes into a single genotype}
\label{tab:genotype_spec}
\small
\begin{tabular}{llll}
\toprule
Gene & Symbol & Type & Domain / constraint \\
\midrule
\multicolumn{4}{l}{\textbf{View builder $a$ (Population $\mathcal{A}$)}} \\
Mask (view 1) & $m^{(1)}$ & binary vector & $\{0,1\}^{d}$, $\lVert m^{(1)}\rVert_0 \ge k_{\min}$ \\
Mask (view 2) & $m^{(2)}$ & binary vector & $\{0,1\}^{d}$, $\lVert m^{(2)}\rVert_0 \ge k_{\min}$ \\
Projection flag (view 1) & $p^{(1)}$ & binary & $\{0,1\}$ \\
Projection flag (view 2) & $p^{(2)}$ & binary & $\{0,1\}$ \\
Projection dim (view 1) & $k^{(1)}$ & integer & $\{2,\dots,k_{\max}\}$ \\
Projection dim (view 2) & $k^{(2)}$ & integer & $\{2,\dots,k_{\max}\}$ \\
Discretization bins (view 1) & $B^{(1)}$ & integer & $\{0,\dots,B_{\max}\}$ \\
Discretization bins (view 2) & $B^{(2)}$ & integer & $\{0,\dots,B_{\max}\}$ \\
\midrule
\multicolumn{4}{l}{\textbf{Policy $b$ (Population $\mathcal{B}$)}} \\
Classifier hyperparameters & $\theta_{\mathrm{clf}}$ & mixed & regularized linear model parameters \\
Initial threshold & $\tau_0$ & real & $(0,1)$ \\
Threshold decay & $\Delta\tau$ & real & $[0,\Delta\tau_{\max}]$ \\
Minimum threshold & $\tau_{\min}$ & real & $(0,1)$, $\tau_{\min}\le\tau_0$ \\
Per-class cap & $q$ & integer & $\{0,\dots,q_{\max}\}$ \\
Margin constraint & $\gamma$ & real & $[0,1]$ \\
Disagreement veto & $\nu$ & binary & $\{0,1\}$ \\
Max iterations & $T$ & integer & $\{1,\dots,T_{\max}\}$ \\
\bottomrule
\end{tabular}
\end{table*}

\section{Genotype encoding}
\label{app:encoding_ssl}
The encoding specifies the view builder component $a$ and the pseudo-labeling policy component $b$, including gene domains and structural constraints. Table~\ref{tab:genotype_spec} summarizes the encoding used in the reported experiments.

\clearpage
\end{document}